\documentclass[useAMS, referee, usenatbib]{biom}

\usepackage{natbib}
\usepackage{amsmath,amsfonts}
\usepackage{graphicx}
\usepackage{subcaption}
\usepackage{enumerate}
\usepackage{natbib}
\usepackage{url} % not crucial - just used below for the URL 
\usepackage{booktabs}
\usepackage{tabularx}
\usepackage{lscape} 
\usepackage{bmpsize}
\usepackage{setspace}
\usepackage{mathrsfs}
\usepackage{amssymb}

\usepackage[hidelinks]{hyperref}
\hypersetup{
    colorlinks,
    linkcolor={red!50!black},
    citecolor={blue!50!black},
    urlcolor={blue!80!black}
}

\usepackage[utf8]{inputenc}
\usepackage{multirow,bigdelim, rotating}
\usepackage{zwcommands}
\usepackage{setspace}
\usepackage{mathtools}

\usepackage{dirtytalk}
\MakeRobust{\say}

\usepackage{xr}

\usepackage{hhline}
\usepackage{natbib}

\title[Dynamic Prediction of Alternating Recurrent Events via Neural Network]{Dynamic Prediction of Alternating Recurrent Events via Neural Network}

\author{Abigail Loe$^{1,*}$\email{aloe@macalester.edu}, 
Susan Murray$^{2}$, and Zhenke Wu$^{2}$\\
$^1$Department of Mathematics, Statistics and Computer Science, Macalester College,\\
Saint Paul, Minnesota, U.S.A.\\
$^2$Department of Biostatistics, University of Michigan, Ann Arbor, Michigan, U.S.A.}
\begin{document}

%  This will produce the submission and review information that appears
%  right after the reference section.  Of course, it will be unknown when
%  you submit your paper, so you can either leave this out or put in 
%  sample dates (these will have no effect on the fate of your paper in the
%  review process!)

% \date{{\it Received October} 2007. {\it Revised February} 2008.  {\it
% Accepted March} 2008.}

%  These options will count the number of pages and provide volume
%  and date information in the upper left hand corner of the top of the 
%  first page as in published papers.  The \pagerange command will only
%  work if you place the command \label{firstpage} near the beginning
%  of the document and \label{lastpage} at the end of the document, as we
%  have done in this template.

%  Again, putting a volume number and date is for your own amusement and
%  has no bearing on what actually happens to your paper!  

\pagerange{\pageref{firstpage}--\pageref{lastpage}} 
\volume{16}
\pubyear{2026}
\artmonth{October}

%  The \doi command is where the DOI for your paper would be placed should it
%  be published.  Again, if you make one up and stick it here, it means 
%  nothing!

% \doi{10.1111/j.1541-0420.2005.00454.x}

%  This label and the label ``lastpage'' are used by the \pagerange
%  command above to give the page range for the article.  You may have 
%  to process the document twice to get this to match up with what you 
%  expect.  When using the referee option, this will not count the pages
%  with tables and figures.  

\label{firstpage}

%  put the summary for your paper here

\begin{abstract}
Alternating recurrent events-- event-times of a specific nature that trigger a secondary refractory period -- occur in a wide-range of fields, including behavioral science, criminal justice, and biostatistics. Analysis of these events requires careful attention to the statistical nuance, including correlated observations and repeated outcomes subject to potential censoring. We develop an online dynamic prediction framework appropriate for predicting subsequent alternating recurrent events, by developing neural network theory for a statistical audiences and applying inverse probability weighted pseudo-observations. The proposed model is applied to dynamically predict alternating recurrent event-free time, showing good performance in simulation, and outstanding capability in application to predicting periods of low mood for first-year medical residents. We close with a discussion.
\end{abstract}

% %%% FOR MHSSS COMPETITION %%%%%
% \begin{abstract}
%     Alternating recurrent events, involving two correlated event types over time, are common in healthcare and behavioral studies. Usually, a primary event triggers treatment and recovery, measured by a secondary time-to-event. Examples include repeated hospitalizations for cancer patients, cycles of addiction and sobriety in alcohol use disorder, or alternating periods of depression and recovery for caregivers. Censoring challenges arise as well as at-risk periods for the primary event are missing if the secondary event has not yet occurred. We introduce a regression framework for censored alternating recurrent events using random forest inverse probability weighting to correct bias in time-to-primary-event analysis caused by informative missingness linked to the secondary event. Our method estimates $\tau$-restricted mean time to the primary event while addressing censoring complexities. Simulations demonstrate strong performance when alternate times are independent or correlated. We apply our approach to mobile health data to assess self-care notification effects on mental states of caregivers for traumatic brain injury patients.
% \end{abstract}

%  Please place your key words in alphabetical order, separated
%  by semicolons, with the first letter of the first word capitalized,
%  and a period at the end of the list.
%

\begin{keywords} Alternating recurrent events; Censored data; Dynamic prediction; Inverse probability weighting; Neural networks; Pseudo-observations.  
\end{keywords}

%  As usual, the \maketitle command creates the title and author/affiliations
%  display 

\maketitle

%  If you are using the referee option, a new page, numbered page 1, will
%  start after the summary and keywords.  The page numbers thus count the
%  number of pages of your manuscript in the preferred submission style.
%  Remember, ``Normally, regular papers exceeding 25 pages and Reader Reaction 
%  papers exceeding 12 pages in (the preferred style) will be returned to 
%  the authors without review. The page limit includes acknowledgements, 
%  references, and appendices, but not tables and figures. The page count does 
%  not include the title page and abstract. A maximum of six (6) tables or 
%  figures combined is often required.''

%  You may now place the substance of your manuscript here.  Please use
%  the \section, \subsection, etc commands as described in the user guide.
%  Please use \label and \ref commands to cross-reference sections, equations,
%  tables, figures, etc.
%
%  Please DO NOT attempt to reformat the style of equation numbering!
%  For that matter, please do not attempt to redefine anything!

\section{Introduction}
\label{s:intro}
Alternating recurrent events are common outcomes in biomedical, sociological, political and behavioral disciplines. For example, periods of hospitalization alternate with periods of discharge. 
Broadly, an alternating recurrent event may be defined as a time-to-event outcome of interest (or \textit{primary event}), which immediately triggers a refractory period (or \textit{secondary event}), wherein an individual is not at-risk for the primary event. Examples may include hospitalization and discharge times as in \citet{huARE}, periods of depressive mood and baseline mood as in \citet{SecondProject}, or recurrent dengue infections, as in \citet{van2025early}.

Accurate and well-calibrated predictions of the next primary alternating recurrent event provide subjects and practitioners with a tool to understand the current health trajectory, as well as informing daily decision making. However, analysis is complicated by the two mutually exclusive event types; for example, when considering time-to next period of depressed mood for medical residents, time spent in a depressed episode competes with time not in a depressed mood. Analyses that fail to account for the time spent in the depressed state, risk biasing the analysis of the time-to the initialization of the next depressed period.
% exacerbation for chronic obstructive pulmonary disease patients, time-exacerbated is relatively short (often patients are discharged from the hospital on the day of exacerbation, and return to normal behaviors), but failing to account for longer secondary states (time-exacerbated) can severely bias analyses of the primary outcome. 

% The majority of literature in the alternating recurrent events space has been semiparametric or parametric in nature. 

There are broadly two schools of thought when approaching recurrent event analysis: (1) state-based modeling using partially observed Markov processes, such as the S(E)IR models in time-series analysis and dynamical systems, where the transition probabilities between states are of interest \citep[e.g.,][]{ionides2006inference,breto2009time}, and (2) semiparametric intensity-based models, where an analyst obtains estimates of covariate effect on relative hazard. This paper focuses on extensions of the semiparametric alternating recurrent events modeling framework to dynamic prediction. To that end, \citet{huARE} extended the Cox-based partial likelihood by modeling hazards of events to obtain interpretable parameter estimates of the intensity of recurrent events. However, their approach does not address potential dependence between recurrent event times, and the resultant, well-known dependent gap-time issue. \citet{lee2019dependence} and \citet{lee2019estimands} chose to model dependent events with a copula approach. \citet{SecondProject} explored a random forest inverse probability weight to de-bias a semiparametric model for alternating recurrent events, handling potential censoring via pseudo-observations. Each of the latter type of model provides, on some level, interpretable parameter estimates, and can inform an analyst of population level dynamics, but requires the specification of a moment-based relationship between covariates and an outcome.

Simultaneously, with the advent of modern computing methods, there has been an increased interest in black-box methods for individual dynamic predictions, which do not require an analyst to specify the any parametric relationship between an outcome and covariates. In the recurrent events case, \cite{DPO_Paper} relied on the presence of smoothers within a semiparametric context to provide updated survival probabilities, while
\citet{LiliBadPaper} and \citet{zhao2020incorporating} constructed a neural network and a random forest, respectively, for dynamic predictions of a single time-to-event. To date, we are unaware of any authors who have proposed dynamic prediction models for alternating recurrent events, particularly in the case of ``online'' or real-time predictions, where predictions for an individual are updated based on newly observed experience. 

In this paper, we present a neural network assisted approach to dynamically predict the time to the primary recurrent event over a $\tau$-length period of follow-up at prespecified check-in time points, borrowing from \citet{TayobMurray:Biostat:2015} to transform the alternating recurrent events into a structure known as a censored longitudinal dataset. This data structure then allows for construction of pseudo-observations to address independent censoring mechanisms \citep{andersen2003generalised, andersen2004regression}. To model these pseudo-observations, we utilize a long short-term memory neural network (LSTM). LSTMs were introduced by \citet{hochreiter1997long} as a solution to the well-documented vanishing gradient problem that plagues recurrent neural networks that use previous outcomes for predictions (for more detail on vanishing gradients, see \citet{hochreiter1997long, hochreiter1996lstm} or \citet{bengio1994learning}). LSTMs use a gating structure to elegantly incorporate previous observed behavior into fitted values at a time (or times) of interest, and can accommodate many different input and output data types. As a subtype of deep learning, LSTMs %The many variations of deep learning (e.g. LSTMs, neural networks with many layers, transformers, generative adversarial networks) 
are widely recognized to excel in settings where there are many covariates with a possibly unknown relationship with an outcome, and prediction, rather than interpretation of coefficients, is of interest. 

By including an inverse probability weight (IPW) in the LSTM loss function, our proposed network accounts for the alternating nature of the data. Rather than providing retrospective (offline), de-biased analysis of a cohort, this paper proposes an online dynamic prediction tool for (primary) alternating recurrent events. We use inverse probability weights (IPW) to correct potential bias, as in \citet{SecondProject}, but compare results in cases where the IPW weight is constructed based on a random forest or an LSTM.
%
% We extend their work by comparing results from a random forest inverse weight and an LSTM-based inverse weight, while also constructing a model for (on-line) dynamic prediction of alternating recurrent events. 
Our weight is then applied to novel dynamic prediction framework using LSTM; however, the field of machine learning and prediction modeling evolves at a fast pace. While our framework is demonstrated by LSTM, it is appropriate for use with the more advanced ML methods plugged-in.
% ***Our IPW weight may be estimated either via an LSTM neural network, or a longitudinal random forest, as described by \citet{hrfRef}.SM: This sentence is important, since it sets up the reader to expect both methods will be looked at.  Change it if you only do one method or the other.  It will be natural to question why you model prediction differently than the inverse weights if you only show random forest weights.***

The rest of this paper is organized as follows. In Section \ref{s:cldf} we describe the censored longitudinal data structure designed for analyzing alternating recurrent events. In Section \ref{s:pseudo}, we create pseudo-observations for addressing independent censoring mechanisms. Sections \ref{s:nnet} and \ref{s:ipwgee} propose two different dynamic prediction models, the former based on a neural network and the latter based on a semiparametric model. The weights used to fit both dynamic prediction models are described in Section \ref{s:estiweight}, while the performance of all models is examined via simulations in Section \ref{s:simulation}. The proposed methods are applied to data arising from the Intern Health Study (IHS) in Section \ref{s:application} to predict the start times of periods of low mood. The paper concludes with a discussion of limitations and future directions in Section \ref{s:discussion}.

\section{Censored Longitudinal Data Structure}
\label{s:cldf}
% be very clear about the logic of how the data is processed adn how the modeling is inserted into the process

% summarize the data strucutre, using different sets of languages. Suppplement can be very helpful here.

%"For the sake of completeness, we are providing all of the notations developed by ..."

% Need to be more concrete, and what in the strucutre is so appealing and a good fit. LSTM is kind of an oldish technique now... Reasons for choosing this model would be helpful.

% it's important to make the distinction between ch2 and what we are doing here. the differences between chapter 2 adn chapter 3 and why i did the things that i did. figuring out why!

For individual $i$, let the alternating recurrent event times $T_{i, 0}^{(1)}< T_{i, 0}^{(2)}< T_{i, 1}^{(1)}<T_{i, 1}^{(2)}<\ldots$ be potentially censored by the random variable $C_i$, where $C_i$ is independent of $T_{i,j}^{(1)}, T_{i,j}^{(2)}$, $i = 1, ..., n$, and $j \geq 0$. Here, $j$ indexes the event pairings, $i$ indexes subjects, and the superscript $k\in \{1, 2\}$ indexes the alternating primary ($k =1$) and secondary ($k=2$) events. In the example from the Intern Health Study (IHS), $T_{i, j}^{(1)}$ marks times when an intern enters low-mood periods. Upon entering the low-mood state at $T_{i,j}^{(1)}$, intern $i$ is no longer at-risk for a low-mood state at $t\in [T_{i,j}^{(1)},T_{i,j}^{(2)})$, until they recover from the low-mood period at time $T_{i,j}^{(2)}$, and re-enter the population of individuals at-risk for periods of low mood. While for our purposes, we have defined $T_{i,j}^{(k)}$ in terms of periods of low and baseline mood, any such alternating structure may be described using these random variables.
%In the example of CareQoL subjects, $T_{i,j}^{(2)}$ would mark the time when the caregiver leaves the depressive state starting at $T_{i,j}^{(1)}$. 
%Recurrent event types are ordered and alternate; thus $T_{i,j}^{(1)}$ is strictly less than $ T_{i,j}^{(2)}$. 

\begin{figure}
    \centering
    \includegraphics[width=\linewidth]{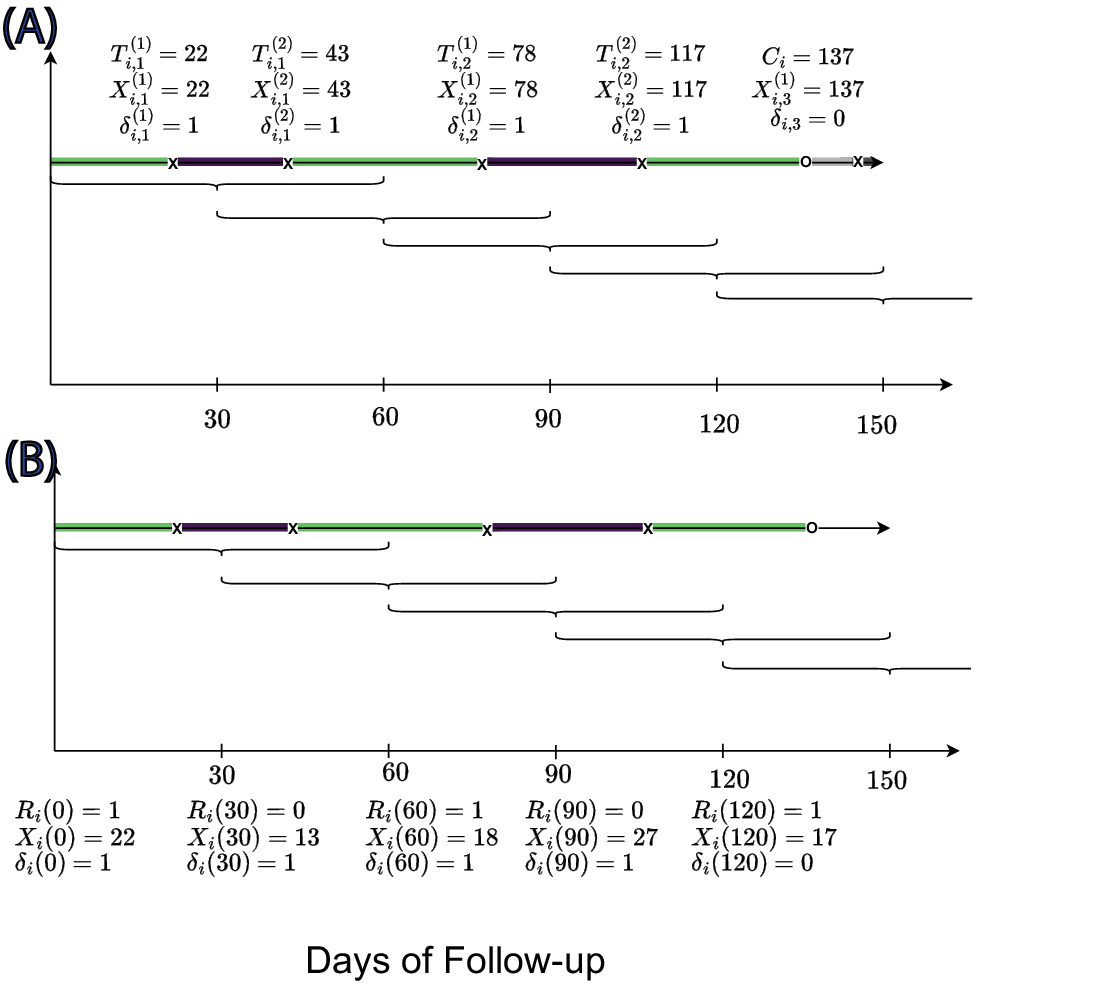}
    \caption{A figure depicting the transformation from alternating recurrent events into the appropriate random variables of the alternating censored longitudinal data structure corresponding to $\mathcal{T} = \{0, 30, 60, 90, 120, 150\}$ and $\tau=60$ days (represented by the curly braces). Panel (A) depicts the traditionally observed alternating recurrent events, with uncensored event times, $T_{i,j}^{(k)}$ listed, with corresponding observed event times $X_{i, j}^{(k)}$ and indicator functions, $\delta_{i,j}^{(k)}$. Periods where subject $i$ is at-risk for a primary alternating recurrent event are depicted in light green, while secondary or recovery states are shown in purple. The $\tau$-length follow-up window corresponding to $\tau=60$ days and $\mathcal{T}_i = \{0, 30, 60, 90, 120, 150\}$ is shown with curly braces. Panel (B) depicts the same $\mathcal{T}_i$, $\tau$, and subject, but records the random variables appropriate for use with the alternating censored longitudinal data structure described in Section \ref{s:cldf}. Potentially time-varying covariates $Z_i(t)$ may be recorded at each $t = \{0, 30, 60, 90, 120, 150\}$.\\
    Alt. text: A visualization of the censored longitudinal dataset. Panel (A) displays traditional survival analysis random variables, while Panel (B) contains the random variables used in the censored longitudinal dataset.}
    \label{fig:cldfpanel}
\end{figure}

For each individual, potential censoring results in observed data $X_{i, j}^{(k)}= \min(T_{i,j}^{(k)}, C_i)$, and censoring indicator, $\delta_{i, j}^{(k)} = I(X_{i, j}^{(k)}< C_i)$, $j=1,\ldots, J_i,k=1,2$. See Figure \ref{fig:cldfpanel}(A) for a depiction of alternating recurrent events, subject to censoring. Rather than analyzing potentially dependent gap times, we follow the approach of \citet{TayobMurray:Biostat:2015}, \citet{Xia:StatMed:2019:regression}, \citet{zhao2020incorporating} and \citet{Chapter1}, and transform longitudinal data into a set of pre-specified follow-up windows of length $\tau$, that start at $t \in \mathcal{T}_i = \{t_0, t_1, \ldots t_{b_i}\}$. 
%If data for subject $i$ are uncensored, then $t_{b_i}$ is selected so that $t_{b_i}+\tau$ does not exceed study follow-up time. In the presence of censoring, $t_{b_i}$ is the last follow-up window start time before $C_i$.
%To construct the censored longitudinal dataset, a user must select an appropriate $\tau$ and $\mathcal{T}$. \cite{XiaMurray:Biostat:2019:commentary} found that in the recurrent event setting, selecting $a = t_k-t_{k-1}$ equal to $\frac{1}{3}$ of the mean gap-time between recurrent events was sufficient to capture approximately 90\% of recurrent events in the original dataset, and $\tau= 2a$ to be the most efficient estimator of $\tau$-RMST. However, to date, more simulations are needed in the alternating recurrent event framework to confirm if this rule of thumb holds. 
% We create the alternating censored longitudinal dataset by the following:
For each $t \in \mathcal{T}_i$, and uncensored individual $i = 1,2, ..., n(t)$ at time $t$, the censored longitudinal data structure maps the original alternating recurrent events, $T_{i, j}^{(1)}< T_{i, j}^{(2)}$ for $j\geq 0$, to the first alternating recurrent event after time $t$. Specifically, we construct the censored longitudinal data for each individual $i = 1, 2, \ldots n(t)$ is:

\begin{itemize}
    \item[1)] Identify the $j$ index of the first alternating recurrent event of type $k = 1,2$ after $t$ by $\eta_i^{(k)}(t)= \min\{j|T_{i,j}^{(k)}> t, j = 0, 1, \ldots \}$;
    \item[2)] Define residual censoring time from time $t$ (the start of the follow-up window) as $C_i(t) = C_i - t$. Let $T_i^{(k)}(t) = T_{i, \eta^{(k)}_{i}(t)}^{(k)}- t$, with $X_i(t) = \min\{T_{i}^{(1)}(t),T_i^{(2)}(t), C_i(t) \}$ measuring time-to-first-observed event time after time $t$ (regardless of its event type), and corresponding censoring indicator, $\delta_i(t) = I\{X_i(t) < C_i(t)\}$; %of any type residual alternating recurrent event-free survival time beyond $t$.% , along with corresponding time-to-censoring , $C_i(t) = C_i - t$.
    %It will be convenient to also define $\tau$-restricted versions, $X_i^\tau(t) = \min(X_i(t), \tau)$, and $\delta_i^\tau(t) = I\{X_i^\tau(t)< C_i(t)\}$.
    \item[3)] Use %$R_i(t) = I\{\eta_i^{(1)}(t) = \eta_i^{(2)}(t)\}$, 
    $R_i(t) = I\{T_{i}^{(1)}(t) < T_i^{(2)}(t)\}$, to indicate that individual $i$ was at-risk for the primary alternating recurrent event as of time $t$. Note that we may modify the definition of $R_i(t)$ by switching the superscripts to shift focus from analyzing the primary event type to analyzing the secondary event type;
    % I[\min\{T_{i}^{(1)}(t), C_i(t)\} < T_i^{(2)}(t)]$ DOES NOT WORK FOR THE CASE WHERE C HAPPENS BEFORE T2 BEFORE T1!!
    
    %(potential) alternating recurrent event after time $t$ was a primary alternating recurrent event. % we need the potential to cover the censored cases!!!
    % \begin{itemize}
    %     \item[a)] If $R_i(t) = 1$, then define residual primary alternating recurrent event-free survival time at $t$ as $T_i(t) = T_{i, \eta_i^{(1)}(t)} - t$. This quantity importantly remains independent of $C_i(t)$. 
    %     \item[b)] For $R_i(t) = 0$, $T_i(t)$ does not exist, or is rather missing due to the secondary alternating recurrent event.
    % \end{itemize}
    \item[4) ] Construct the censored longitudinal dataset at $t$ by $\big\{X_i(t), R_i(t), \delta_i(t), \bZ_i(t),  i = 1, 2, \ldots n(t) \big \}$ where $\bZ_i(t)$ are observed covariates as of time $t$. Figure \ref{fig:cldfpanel}(B) displays the censored longitudinal data tuples (except the covariates), $\{R_i(t), X_i(t), \delta_i(t)\}$, that correspond to the censored alternating recurrent events depicted in Figure \ref{fig:cldfpanel}(A).
\end{itemize}

% the 1-R-i(t) term, we are not seeing that piece.

In this paper, let $T_i(t)$ be the time to primary type of recurrent event after time $t$. However, for any subject with $R_i(t) = 0$, $T_i(t)$ is unobserved, because at time $t$ this subject has recently experienced the primary event, and has entered into a period anticipating the secondary, not the primary, event type. For example, in the IHS, $T_i(t)$ may be defined as time-to-low-mood after time $t$, but for subjects who are already depressed in mood at time $t$ and waiting to recover, we do not observe $T_i(t)$.
%That is, $T_i(t) = R_i(t) \times T_i^{(1)}(t)$, as it will be convenient to define $T_i(t)=0$ when subject $i$ is not in the at-risk set. **** This definition is not describing the ideal $T_i(t)$, the $\{1-R_i(t)\}\times T_i^{(1)}(t)$ piece.****
%Let $n_R(t)=\sum_{i=1}^{n(t)}R_i(t)$ be the number of uncensored subjects at-risk for the primary recurrent event at time $t$, with complete data on the primary recurrent event endpoint.
We will hereafter refer to $\mathcal{O}(t)=\{ i = 1, 2, \ldots, n_R(t): R_i(t) = 1\}$ as the set of individuals at time $t$ with observed $T_i^{(1)}(t)$ and who have not reached a censoring event, with $|\mathcal{O}(t) | = n_O(t)\leq n(t)$. % being the number of individuals in the set $\mathcal{O}(t)$. 
We define % = \sum_{i \in \mathcal{CC}(t)}R_i(t)$
$\{X_i(t), \delta_i(t): i \in \mathcal{O}(t)\}$ as the complete case (non-missing) data for the primary recurrent event of interest at $t$. 
In Sections \ref{s:nnet} and \ref{s:ipwgee}, we propose models for dynamically predicting $\tau$-restricted time-to-the-primary-event $\min\{T_i(t),\tau\}$ using $E[\min \{T_i(t), \tau \}]$. 
%The $\tau$-restricted mean provides expected survival time within a clinically relevant threshold an aid for decision making, and has gained popularity in recent years \citep[e.g., ][]{royston2013restricted, OG_rmst,  RMST_for_clinical_Results, rmst_vs_conventional}. 
In particular, we introduce (weighted) LSTM networks and weighted semiparametric models fit via generalized estimating equations to ensure that primary event analyses account for missing subjects due to the secondary state. Prior to providing model specifications in Sections \ref{s:nnet} and \ref{s:ipwgee}, we discuss a pseudo-observation approach to account for censoring by $C_i(t)$.

% We now formally express our parameter of interest as $\tau$-restricted mean primary alternating recurrent event free survival time at time $t$, or $\min(T_i(t), \tau)$. 
%The resultant longitudinal nature of our data requires a method sensitive to the correlation within subject, and measures repeated over time.

\section{Preliminary: Pseudo-observations}
\label{s:pseudo}
Here, we provide a primer on the construction of weighted pseudo-observations, a variation of jack-knife methodology originally proposed by \citet{andersen2003generalised} and \citet{andersen2004regression}. Pseudo-observations are typically constructed by selecting an unbiased estimator for an estimand of interest, i.e., for our dynamic prediction model, $\tau$-restricted mean survival after time $t$, or $E[\min \{ T_i(t), \tau\}]$. The estimator is then constructed from two different data subsets, one with all observations in $\mathcal{O}(t)$, and another of $\mathcal{O}(t)$ with the $i$th observation omitted. 

Let $S^{\mathcal{O}(t)}(u) =P(T_i^{(1)}(t) > u) $ be the survival curve as a function of $u$ corresponding for subjects at-risk for the primary event at time $t$, and $\theta(t) = E[\min\{T_i^{(1)}(t), \tau\}|R_i(t) = 1] = \int_0^\tau S^{\mathcal{O}(t)}(u) du $. $\theta(t)$ has a consistent nonparametric estimator $\hat \theta(t)= \int_0^\tau \hat{S}^{\mathcal{O}(t)}(u) du$, where $\hat{S}^{\mathcal{O}(t)}$ denotes the Kaplan-Meier estimator for the survival curve using subjects in $\mathcal{O}(t)$. Similarly, one can construct $\hat \theta^{(-i)}(t)$, by using the Kaplan-Meier survival function estimated from $\mathcal{O}(t)$ without subject $i$. Taken together, a pseudo-observation for $E[\min\{T^{(1)}_i(t), \tau\}]$ for subject $i$ at time $t$ is defined as $$PO_i^{\tau}(t) = n_O(t) \hat{\theta}(t) - \{n_O(t) -1\}\hat{\theta}^{(-i)}(t).$$ \citet{andersen2003generalised} showed that in regression settings, the univariate pseudo-observation may replace the bivariate censored survival outcome; for longitudinal settings, results pertaining to estimating equations from \citet{robinsrotnitzky1995analysis} may be applied. These pseudo-observations are calculated using the cohort of individuals in $\mathcal{O}(t)$; details on how to correct the pseudo-observations during estimation to account for selection into $\mathcal{O}(t)$ are given in Section \ref{s:estiweight}.

\section{Statistical Review of Neural Networks}
\label{s:nnet}
% REALLY HELPFUL NNET DOCUMENTATION from a coding perspectiv: https://medium.com/analytics-vidhya/lstms-explained-a-complete-technically-accurate-conceptual-guide-with-keras-2a650327e8f2
Neural networks are a popular way to model potentially complex functions with many variations and adaptations
%.  At their simplest, a neural network approximates a linear regression (see Figure \ref{fig:basicnet}), though many variations and adaptations of neural networks exist 
\citep[see, for example][]{ LSTMlanguage, LSTMsleep, LSTMfuelcells}.
% We apply long short-term memory networks for two purposes, the first to fit a model estimating $P\{R_i(t)|W_i(t)\}$, and the second to dynamically predict $E[\min\{T_i(t), \tau\}]$.
In this section, we describe the neural network loss function with corresponding architecture appropriate for online dynamic prediction of the censored longitudinal pseudo-observation dataset. An introductory primer on the use of neural networks may be found in \citet{loe2026neuralnetworkslinearregression}, while additional practical considerations of applying this neural network algorithm in our setting are given 
%Further details of , and detail potential inclusion of hyperparameters 
in Section \ref{s:LSTMspecifics}.

For dynamic online prediction of the $\tau-$restricted mean for the follow-up window that starts at $t_k, t_k \in \mathcal{T}$, or $E[\min\{T_i(t_k), \tau\}\mid \Bar{\bZ}_i(t_k)]$, where $\bar{\bZ}_i(t)$ is the vector of covariates values $\bZ_i(t_{k}), \bZ_i(t_{k-1})\ldots \bZ_i(t_0)$, we follow the neural network architecture laid out in Figure \ref{fig:lstm}B. This architecture is commonly termed long short-term memory networks, or LSTM, and is reviewed in \citet{loe2026neuralnetworkslinearregression}. As the field of machine learning progresses, this stage of our analysis approach can be replaced with more modern versions of deep learners, such as transformers or gated recurrent units. For dynamic prediction of $E[\min\{T_i(t_k), \tau\}\mid\Bar{\bZ}_i(t_k)]$, the selected super-architecture only connects architectures across previous follow-up windows.
% indexed by $t_0,\cdots, t_{k-1}$. 
That is, we select
$\beta_{ \cdot, t}, \alpha_{\cdot, t}, t < t_k - \tau $ that minimize:
$$
\arg \min \limits_{\beta_{ \cdot, t}, \alpha_{\cdot, t}} \sum_{t< t_k-\tau } \{\hat{\mathbf{PO}}^\tau (t) - g^{-1}(\eta_t)\}^\top \mathbf{W}(t)\{\hat{\mathbf{PO}}^\tau (t) - g^{-1}(\eta_t)\},
$$
where $\mathbf{W}(t)$ is a $n(t)\times n(t)$ diagonal matrix, with diagonal entries $\frac{R_i(t)}{\hat{P}\{R_i(t) = 1 \mid \Bar{\bZ}_i(t)\}}$, and zeros off the diagonal. 
Components of $\eta_{t_k}$ are laid out in Figure \ref{fig:lstm}B. 
For estimation of $E[\min\{T_i(t_k), \tau\}\mid \Bar{\bZ}_i(t_k)]$, the final activation link, $g,$ is taken to be the identity link. Estimation of $\hat{P}\{R_i(t) = 1 \mid \Bar{\bZ}_i(t)\}$ is covered in Section \ref{s:estiweight}.

For details regarding our selection of training, validation and testing cohorts via the rolling-origin algorithm \citep{armstrong1972comparative}, see Supplemental Materials Section \ref{s:LSTMspecifics}.

\begin{figure}
    \centering
    \includegraphics[width =\textwidth]{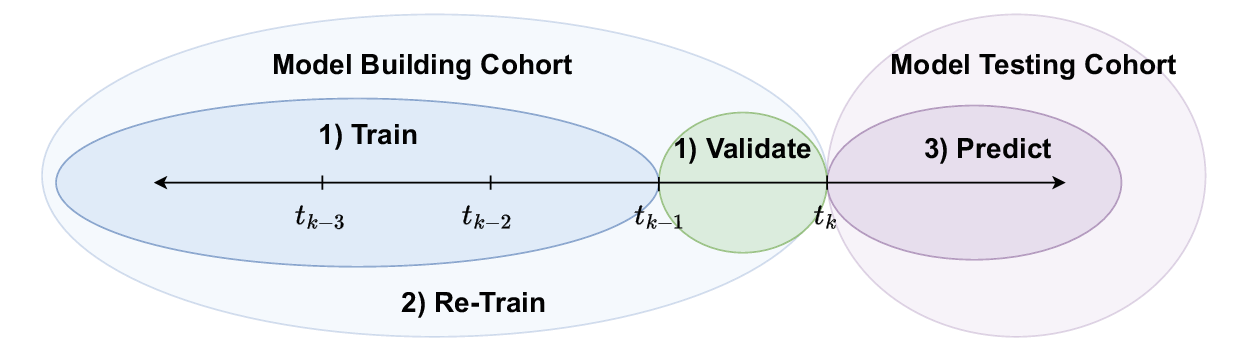}
    \caption{A figure depicting the training, validation, and testing split for the rolling-origin algorithm, mapped onto the censored longitudinal data structure. The model building component contains data from $\{t_0, t_1, \ldots t_{k-1}\}$, while model testing comes from data observed at $t_k$. Within the model building cohort, the network is first trained on data from $\{t_0, t_1, \ldots t_{k-2}\}$, while data observed in the window starting at time $t_{k-1}$ are used to validate. When the optimal number of epochs has been reached, $n_{opt}$, initial training stops. Re-train the model using data observed at window start times $\{t_0, t_1, \ldots t_{k-1}\}$. This model is then tested for data at $t_k$, to obtain dynamic predictions of $E[\min\{T_i(t_k), \tau\}]$.\\
    Alt. text: A depiction of the rolling-origin training, validation and testing split for predicting at time $t_k$. Various data cohorts are displayed in different colors, while the steps in obtaining predictions are labeled ``1)," ``2),", and ``3)."}
    \label{fig:rolling_origin}
\end{figure}

\section{Weighted Semi-parametric Dynamic Prediction Model}
\label{s:ipwgee}
At each $t_k \in \mathcal{T}$, we may fit $E[\min\{T(t_k) , \tau\}|\Bar{\bZ}_{t_k}] = \beta^\top \bZ_{t_k}$ via  generalized estimating equations. As this paper is concerned with real-time dynamic predictions, at each $t_k$, we select data corresponding to $t < t_k-\tau$ for estimating $\hat{\beta}$, and data at $t_k$ to obtain predictions. Similar to the LSTM training, testing and validation algorithm described in Section \ref{s:nnet:lstmDesc}, we need a model warm-up period while waiting for complete follow-up in the first $\tau$-length window. Point estimates are then the solution to: 
$$
\mathbf{0} = \sum_{i =1}^{n} \Bar{\bZ}_i \bW_i \mathbf{\Lambda}_i^{-1}(\hat{\mathbf{PO}}_i^\tau - \Bar{\bZ}_i^\top \beta)
$$
where $\bW_i$ is a diagonal matrix (which weights least squares loss) with entries $\frac{R_i(t)}{\hat{P}\{R_i(t) = 1\mid \Bar{\bZ}_i(t)\}}$, for $t = t_0, t_1 \ldots t_{k-1} < t_k -\tau$ and $\mathbf{\Lambda}_i$ is the usual matrix product of variance functions and the working correlation structure. Estimates from this method will be termed ``online GEE," in reference to the suitability for online, or real-time, predictions. Parameter estimates may then be applied to the cohort of data observed at $t_k$.
Estimates from this model have the benefit of interpretability, and do not suffer from the same concerns surrounding tuning and convergence in deep learning approaches. However, in the presence of highly correlated covariates, such as the useful history covariates of \citet{Chapter1}, a semiparametric approach may not be fully estimable, particularly for elements of the variance-covariance matrix. %Yet even in the setting where covariates are highly correlated, and model identifiability is questionable, the linear predictor obtained via GEE is still reliable for prediction. However, semi-parametric models of this nature rely on correctly specifying the relationship between covariates and outcome. % Moreover, semiparametric models rely on the correct specification of covariates and outcome.

\section{Estimation of Inverse Probability Weights}
\label{s:estiweight}
The pseudo-observations from Section \ref{s:pseudo} are constructed only using the set of individuals in $\mathcal{O}(t)$, yet  \citet{SecondProject} showed that by weighting pseudo-observations with inverse probabilities, %one may obtain correctly specified regression models that unbiasedly estimate the restricted mean. In this manner, 
an analyst may account for non-random selection into the at-risk set. The inverse probability weights may either come from an LSTM, as described in Section \ref{s:nnet}, or a longitudinal random forest, as proposed by \citet{hrfRef}. In estimating $\hat{P} \{R_i(t) = 1 |\Bar{\bZ}_i(t)\}$ via LSTM, we assume the network depicted in Figure \ref{fig:lstm} has a logistic link function $g$, and select $\beta_{ \cdot, t}, \alpha_{\cdot, t}$ that minimize: %That is, for epoch $(a)$ of the network at time $t_k$, 
% \begin{align*}
%     \hat{P}^{(a)}\{R_i(t_k) = 1 \mid Z(t_k)\}= \hat{\pi}^{(a)}_i(t_k)  = \text{expit} \bigg \{ o^{(a)}_{t_k} \tanh\big(f^{(a)}_{t_k}\cdot  C^{(a)}_{t_{k-1}} + i_{t_k} \cdot C^{*(a)}_{t_k} \big )\bigg \}.
% \end{align*} 
$$
\arg \min \limits_{\beta_{ \cdot, t_k}, \alpha_{\cdot, t_k}} \sum_{t\leq t_k} \left[{\mathbf{R}}(t)\cdot \log \{g^{-1}(\mathbf{\eta}_t)\}\right] + \left[\{\mathbf{1} - \mathbf{R}(t)\}\cdot \log \{\mathbf{1} - g^{-1}(\mathbf{\eta}_t)\}\right].
$$
This network is trained using observed values $R_i(t)$, $i =1, \ldots n(t)$ for all time points up to and including $t_k \in \mathcal{T}$. We perform a 2-fold validation/training algorithm for selecting number of epochs, and obtain fit values from the validation set. Specifically, for each $t_k \in \mathcal{T}$, we split the data into two equal sized training and validation cohorts, based on subject IDs, and including all time points $t\leq t_k$. Training on a given learning rate, $\triangledown_m$ proceeds with the training set, until validation loss fails to decrease for $\max_p$ epochs. The training model corresponding to the best validation loss is restored, and $\hat{P}\{R_i(t_k) = 1 | \bar{Z}_i(t_k) \}$ is calculated for the validation cohort via the trained model. The procedure then repeats, with the cohort labels reversed. Because this is a cross-validation algorithm, and does not rely on the rolling-origin cross validation algorithm from Section \ref{s:nnet}, the model warm-up set is simply $\{t_0\}$; thus for values $t_1, \ldots, t_b$, one may obtain LSTM weights in this manner.

\begin{figure}
    \centering
    \includegraphics[width=\linewidth]{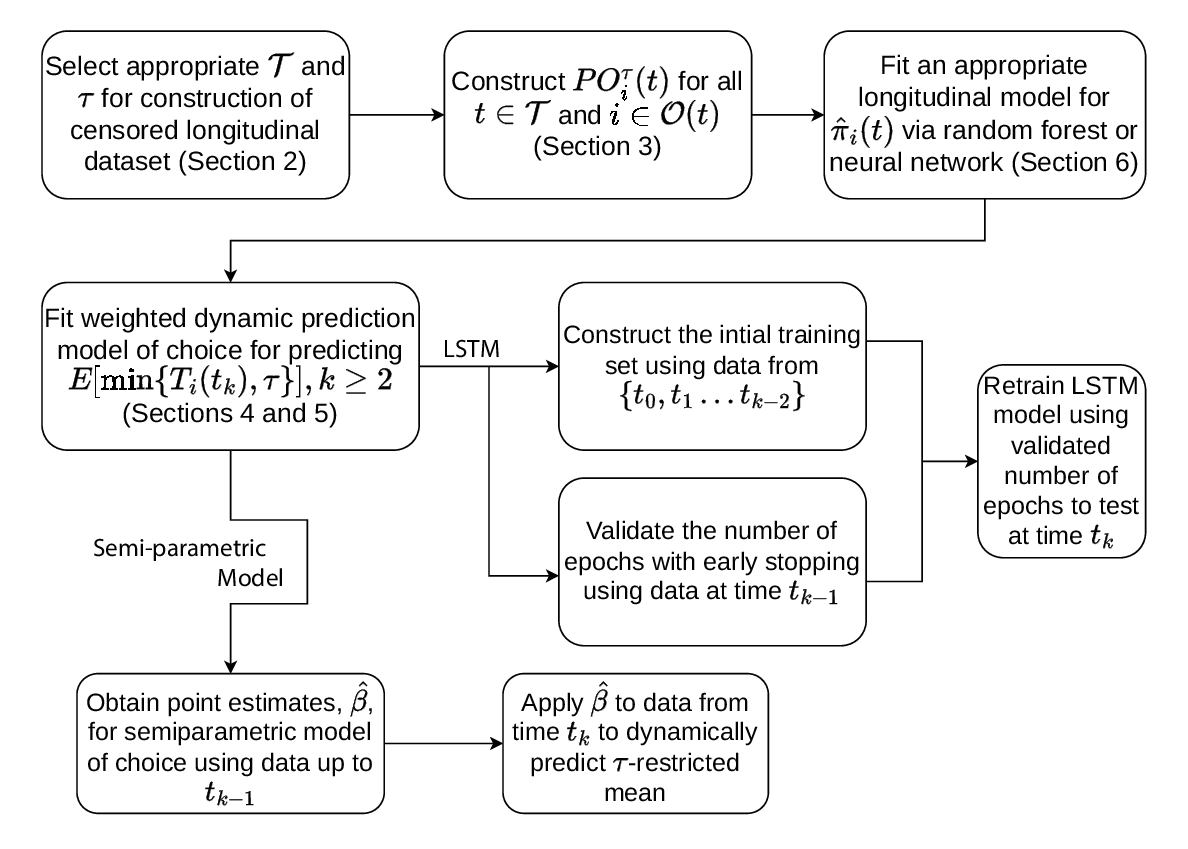}
    \caption{A figure displaying the analysis pipeline for on-line (real-time) prediction using random forest or LSTM IPW weights and the rolling-origin training and validation split for both the LSTM and GEE-based models.\\
    Alt text: A flowchart displaying steps in the analysis pipeline, starting at construction of the censored longitudinal dataset, and ending with the acquisition of dynamic predictions.}
    \label{fig:flowchart}
\end{figure}

In this paper, we compare the performance of weights estimated via LSTM (as described above) and a longitudinal random forest. Random forests have many of the same desirable properties as neural networks; they can fit a model to potentially correlated covariates, as well as non-linear relationships between covariates and outcomes. Additionally, random forests may be programmed to always incorporate certain predictors into their model, providing the analyst with a greater degree of control than a more black-box neural network. Generally, they have fewer hyperparameters to tune (the number of trees, the minimum number of observations in a node, and the maximum number of splits within a tree), and rather than use a training and testing cohort, random forests use a strategy known as ``bagging," which naturally creates distinct cohorts of data for training and testing (for further detail, see \citet{SecondProject}). We apply the longitudinal random forest for dynamic prediction of at-risk status. A flowchart depicting the entire analysis pipeline is shown in Figure \ref{fig:flowchart}.

\section{Simulation Study}
\label{s:simulation}
%As detailed in Section \ref{s:intro}, neural networks (and other machine learning approaches) are known  for fitting black box models with an unknown relationship between covariates and outcomes. 
The goals of our simulation study are three-fold: (1) to examine the one-step online predictive ability of the proposed LSTM model for the primary recurrent event of interest when both primary and secondary events have non-trivial associations with covariates, (2) to describe pragmatic approaches to including history covariates in the algorithms, and (3) to evaluate the performance of a longitudinal random forest inverse weight compared to an LSTM inverse weight.

We first describe the generation of covariates, $\bZ$ that are used in generated correlated gap times between alternating recurrent evens, $G_{i, j}^{(1)} = T_{i,j}^{(1)}- T_{i, j-1}^{(2)}$ and $G_{i,j}^{(2)} = T_{i,j}^{(2)}- T_{i, j}^{(1)}$. We draw
$Z_1 \sim \text{Cauchy}(\theta = 4, \sigma = \frac{1}{100})$, $Z_2 \sim \Gamma(3, \frac{1}{3}), Z_3\sim \text{Beta}(5, 1)$, $Z_4\sim \text{Exponential}(1.7)$, $Z_5, Z_6 \sim \text{MVN}([13, -.5]^\top, \big(\begin{smallmatrix}
    1 & .9\\
    .9 & 1
\end{smallmatrix}\big) )$, $Z_7$ through $Z_{100}$ are random normal noise covariates, with $\mu = \text{Unif}[-1, 1]$ and $\sigma^2 =1 $, $Z_{101}\ldots Z_{124}$ are multivariate normal, where $Z_{101}$ to $Z_{108}$ have mean $0.3$, $Z_{109}$ to $Z_{116}$ have mean $0.1$ and $Z_{117}$ to $Z_{124}$ have mean $-0.3$, each with variance $ 0.4$ and correlation $ 0.28$. 

To obtain correlated gap times, we draw from a multivariate normal distribution with $\mathbf{\mu}_i = (\mu_i^{(1)}, \mu_i^{(2)}, \mu_i^{(1)},\mu_i^{(2)}, \ldots )$ and variance-covariance matrix $\Sigma$, which has a block diagonal form, with entries in each block: 
\begin{equation*}
    a_{j,k} = \begin{cases}
        \rho^{(1, 1)}\sigma^2 & j \neq k, j \text{ and } k \text{ are both odd}\\
        \rho^{(2, 2)}\sigma^2 & j \neq k, j \text{ and } k \text{ are both even}\\
        \rho^{(1, 2)}\sigma^2 & j \neq k, j \text{ and } k \text{ have different parity}\\
        \sigma^2 & j = k
    \end{cases}
\end{equation*}
Here, $\rho^{(k, \ell)}$ is the correlation between $G_{i, j}^{(k)}, G_{i, j}^{(\ell)}$. We select $\rho^{(1, 1)} = 0.3$,  $\rho^{(2, 2)} = 0.2$, $\rho^{(1,2)} = -0.2$, and $\sigma^2=10^{-5}$. We defined $\mu_i^{(1)} = \exp( -\sin{Z_1} + 2 Z_2 Z_3+ (Z_2 Z_4^2)^{\frac{1}{3}} - 3\sqrt{Z_4}+ \cos(Z_5) Z_6^2 + \mid Z_6\mid + 2\sin\{\cos(Z_4)\}- 2\tan(Z_3))$, and $\mu_i^{(2)} = \exp(Z_{101} + \ldots Z_{124}) Z_3 (\Tilde{X}_6 + |\Tilde{X}_6+.3|)$, where $\Tilde{X}_6$ is centered and scaled $Z_6$.
Summing across gap-times yields alternating recurrent event time $T_{i,1}^{(1)}< T_{i, 1}^{(2)}< \ldots$, and we follow the advice of \citet{Xia:StatMed:2019:regression} and apply a burn-in period of at least $b_{pre} = 
\max \frac{5}{\lambda_i}$ duration prior to defining the analysis dataset. Combinations of covariates that yield $\mu_i^{(1)}$ values outside of $0.45$ and $2$ and $\mu_i^{(2)}$ outside of $0.2$ to $0.6$ are discarded. %The large number of noise covariates are expected to be especially challenging for the LSTM algorithm that is already subject to identifiability issues. 
The censored longitudinal data structure takes the form $\{0, .5, 1, \ldots, 11.5\}$ months, with $\tau = 0.5$ months and $C_i - 2 \sim \text{Exponential}(\frac{1}{20})$ to mimic a year of follow-up with approximately 40\% of observations censored.

% In each simulation, we fit two different methods for calculating the inverse probability weights, the first being an LSTM, with the second a longitudinal random forest.  %***Need some clarity about run-in windows versus prediction windows here as well***
% For all subjects, $R_i(t_0) =1$ by definition. At $t = t_1, t_2\ldots t_b$, we fit $P\{R_i(t)=1|\Bar{Z}_i(t)\}$ via two-fold cross validation to select optimal epochs and learning rate for the LSTM weights. Details on the hyperparameters selected may be found in Supplementary Materials, Section \ref{s:supp:ipw.lstm}.

As the goal of this paper is to dynamically model $E[\min\{T_i(t), \tau\}]$, we consider two settings of history covariates proposed by \citet{Chapter1} and extended in \citet{SecondProject}. The first setting is where subjects have an exact recollection of their last pair of alternating recurrent events within the previous year, termed $\bH_h$ for ``high-quality history," and the second set of history covariates, $\bH_\ell$ for ``low-quality history," where no history prior to study time is recorded, though once follow-up commences, these covariates may be recorded. %\citep{Chapter1}. 
In this latter case, \citet{Chapter1} proposed a multiple imputation approach for handling missing history covariates. Here, we describe results using a pragmatic approach that incorporates a single imputation of the history covariates at $t_0$ from the average patient population. For later follow-up windows with sufficient observed patient history for each individual, this $t_0$ impute is dropped from patient history covariate calculations.
The various algorithms include history covariates along with other covariates in estimating inverse probability weights as well as modeling $E[\min\{T_i(t), \tau\}]$; GEE models include history covariates as main effects. 
%These covariates are used in estimating inverse probability weights, and the effect of their presence or absence on the proposed dynamic prediction models is evaluated. For GEE based models, history covariates are incorporated as main effects. %If the inclusion of history covariates leads to model instability due to correlated covariates (particularly at early time points), or inestimable variance-covariance matrix components, history covariates are excluded from modeling. On the other hand, LSTM models flexibly incorporate history covariates at all time points, as concerns relating to highly correlated covariates, interactions, and Gram matrix inversion are not applicable to the LSTM network proposed.

Neural network inverse weights are estimated with an LSTM architecture with $1$ dimensional subject embeddings, $8$ dimensional time embeddings, $1$ layer, $\eta\in \mathbb{R}^{n(t)\times 32}$ and $p_d=0$. Hyperparameters for the inverse weight LSTM were selected via a grid search. Neural network architecture for dynamic predictions of the $\tau$-restricted mean was also selected using a rolling-origin grid-search; based on the types of covariates available ($\bH_\ell$ or $\bH_h$) as well as inverse weighting strategy (LSTM or RF), different architectures yielded lower testing losses. Exact hyperparameter specifications are shown in Table \ref{tab:hyperparams}. Each network has 1 dimension for subject embeddings, and 8 dimensions for time embeddings. The grid search prioritizes deeper architectures (larger $h$) for LSTM weights, and more dropout for $\bH_h$ settings.

% For neural network weights with $\bH_\ell$ covariates, we select 1 dimension for subject and $8$ dimensions for time embeddings, hidden dimension $h=8$, with 1 layer and $p_d=0$ (no parameters randomly set to zero). For $\bH_\ell$ covariates with RF weights, subject embedding has dimension 1, time embedding has dimension 8, $h = 4$, with one LSTM layer, and $p_d = 0$. For $\bH_h$ covariates with neural network weights, subject and time embedding dimension remains $1$ and $8$ respectively, while $h$ is increased to 64, with two LSTM layers, and $p_d=0.2$. For $\bH_h$ covariates with RF weights, subject embedding has dimension 8, time embedding has dimension 4, $h = 64$, with two LSTM layers, and $p_d = 0.1$. As a result, restricted mean LSTM models in the same history covariate setting correspond to distinct architectures; specifically, the $\bH_\ell$ LSTM with neural network weights is more complex than the $\bH_\ell$ LSTM with RF weights, and the $\bH_h$ LSTM with neural network weights is less complex with more regularization than the $\bH_h$ LSTM with RF weights.

We select $n_{opt}, \triangledown_{opt}$ from candidates $1, \ldots 250$ and $\{0.005, 0.001\}$, respectively, using the rolling-origin algorithm from Section \ref{s:nnet}. We compare this LSTM to two different online semiparametric GEE models. The first model is
%reflects the generating hazard of the primary recurrent event, modeling 
$E[\min\{T_i(t), \tau\}\mid \bZ_i(t)] = \beta^\top \mathbf{Z}_i(t)$, where $\mathbf{Z}_i(t)$ includes covariates $\bZ_1$ through $\bZ_{6}$ in the correct functional form used for modeling $\mu_i^{(1)}$ as well as history covariates. We term this model ``GEE Oracle," though exact oracle modeling of $T_i(t)$ would include terms that count number of observed events (for further details, see Section \ref{s:supp:oracle}). The second online GEE model, or ``GEE Main Effects," considers $E[\min\{T_i(t), \tau\}\mid \bZ_i(t)] = \beta^\top \bZ_i(t$) where $\bZ_i(t)$ contains $\bZ_{1}, \ldots \bZ_{124}$ as well as history covariates as main effects.

\subsection{Evaluation Metrics}\label{s:simulation:eval}
Model performance is evaluated in the testing set using average bias, the C-statistic (also referred to as the index of concordance or Harrell's C-index)
%, a nonparametric measure of concordance 
\citep{HarrellsCStat} and root error sum of squares, $\sqrt{\frac{1}{n_\mathcal{O}(t)}\sum_{i \in \mathcal{O}(t)} [\hat{E}[\min\{T_i(t), \tau\}|\Bar{\bZ}_i(t)\} - \hat{PO}_i^\tau (t)]^2}$, denoted ``Testing RESS." C-statistics measure predictive discrimination, and a value of approximately $0.50$ indicates poor discriminative ability, while perfect prediction yields a C-statistic of $1$. Testing RESS is the root of the average of error sum of squares; while it is not the theoretical function of expectation $\sqrt{E_{T_i(t)}[\{T_i(t)-\hat{PO}^\tau_i(t)\}^2]}$, or RMSE, it is an observed average that approximates RMSE. As an estimator of RMSE, RESS provides information about both bias and variance of predictions in the held-out testing set.
%Briefly, the C-statistic is calculated by comparing the ordering of any two fitted risk scores and their true values. That is, if $\hat{y}_i(t) < \hat{y}_j(t)$, and $y_i(t) < y_j(t)$, then the pairing is termed ``concordant." If the pairings do not agree, the pairing is known as ``discordant." The C-statistic is the number of concordant pairs divided by the number of total comparisons. 
 %defined as $\sqrt{\frac{1}{n_\mathcal{O}(t)}\sum_{i \in \mathcal{O}(t)} [\hat{E}[\min\{T_i(t), \tau\}|\Bar{\bZ}_i(t)\} - \hat{PO}_i^\tau (t)]^2}.$

The true value of $E[\min\{T_i(t), \tau\}\mid \bZ_i(t)]$ needed to define bias is determined upon noting that the conditional distribution of $T_i(t)$ given $\bZ_i(t)$ is
%through $\mu_i^{(k)}$ so that:
$
%T_i(t) \sim 
\text{Normal}\big[ \{N_i^{(1)}(t^-)+1 \}\mu_i^{(1)}+N_i^{(2)}(t^-)\mu_i^{(2)} - (t+b_{pre}), \Tilde{\sigma}^2 \big ],
$
where $N_i^{(k)}(t^-)= \sum_j I(T_{i,j}^{(k)}<t + b_{pre})$ is the number of events of type $k$  experienced by subject $i$ prior to time $t$ including those accumulated during the burn-in period of duration, $b_{pre}$. $\Tilde{\sigma}^2 =  \sum_{1\leq j \leq \{N_i^{(1)}(t^-)+ N_i^{(2)}(t^-)+1\}} \sum_{1\leq k \leq \{N_i^{(1)}(t^-)+ N_i^{(2)}(t^-)+1\}} a_{j,k}$, where $a_{j,k}$ is the $j$, $k$ entry of $\Sigma$ defined above.  
%Our simulation setting assumes the following relationship between covariates and a $\tau-$ restricted mean survival time:
From this we use the relationship,
$$
E[\min\{T_i(t), \tau\}\mid \bZ_i(t)] = \int_{-\infty}^{\tau^-} y f_{T_i(t)}(y) dy + \tau \{1-F_{T_i(t)}(\tau)\},
$$
where $f_{T_{i}(t)}$ and $F_{T_{i}(t)}$ are the probability distribution functions and cumulative distribution functions of the time-to-first primary alternating recurrent event at time $t$, $T_i(t)$. For further detail, see Supplemental Materials, \ref{s:supp:oracle}.

\subsection{Results}
\begin{figure}[htbp]
    \centering
        \includegraphics[ width=\textwidth]{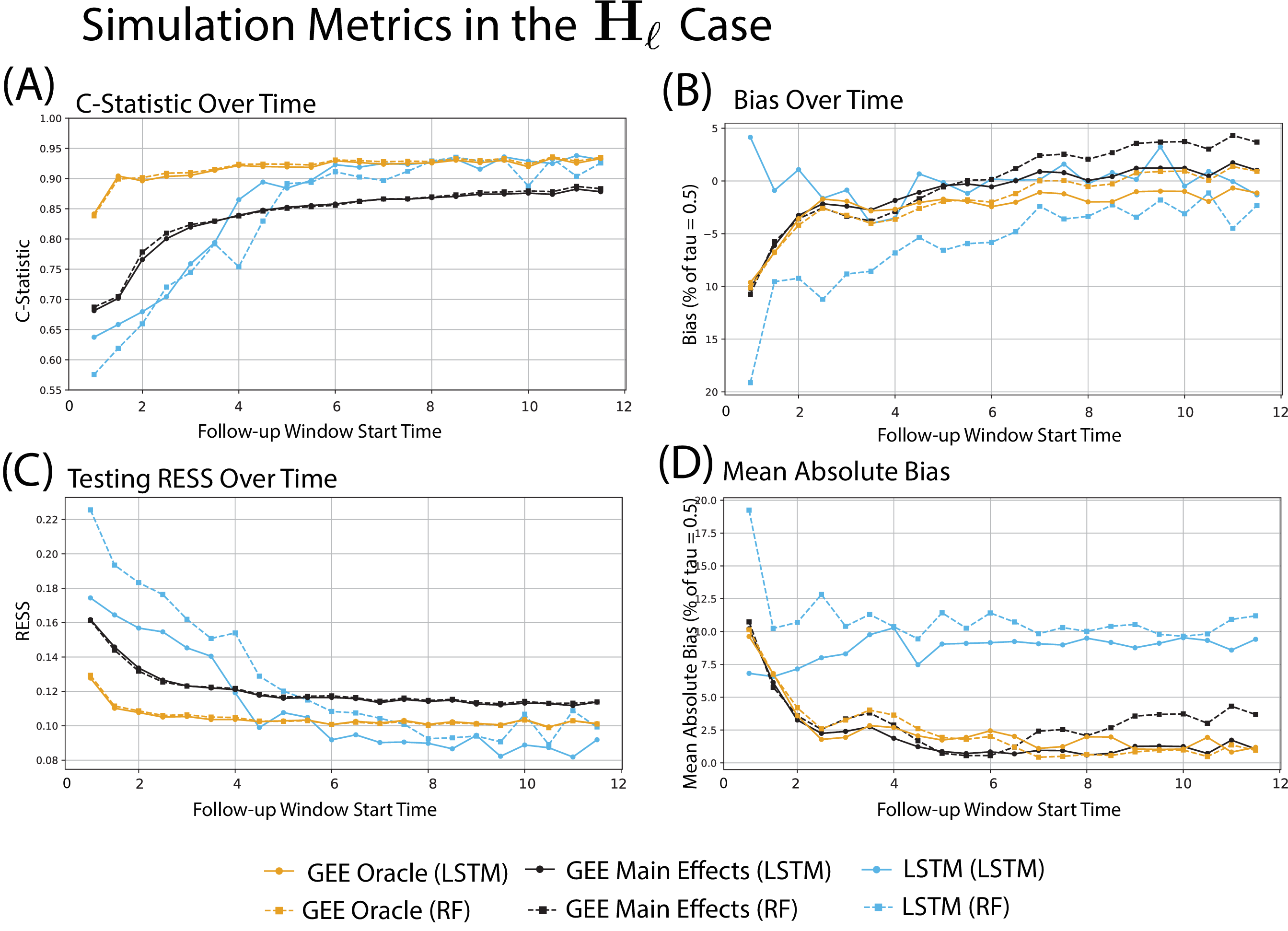}
\caption{A panel graphic displaying the performance of the dynamic prediction models under consideration in Section \ref{s:simulation} for the $\bH_l$ setting. Counter clockwise from upper left is (A) c-statistic over time, (B) percent bias (in terms of $\tau$) over time, (C) testing root mean squared error over time, and (D) $n_{opt}$ over time. The method used to predict $E[\min\{T_i(t), \tau\}] = \Bar{\bZ}^\top \beta $ is listed first (yellow for GEE Oracle, black for GEE Main Effects, and blue for LSTM), while the weight model is listed in parenthetical (with LSTM weights graphed in a solid line, while random forest weights are dashed). All results are from $200$ replicates with $n = 2500$ each.\\
Alt text: A panel graphic displaying simulation metrics for the dynamic prediction models in the low-quality history information setting.
} \label{fig:only_hist_metrics}
\end{figure}

\begin{figure}
    \centering
    \includegraphics[width=\textwidth]{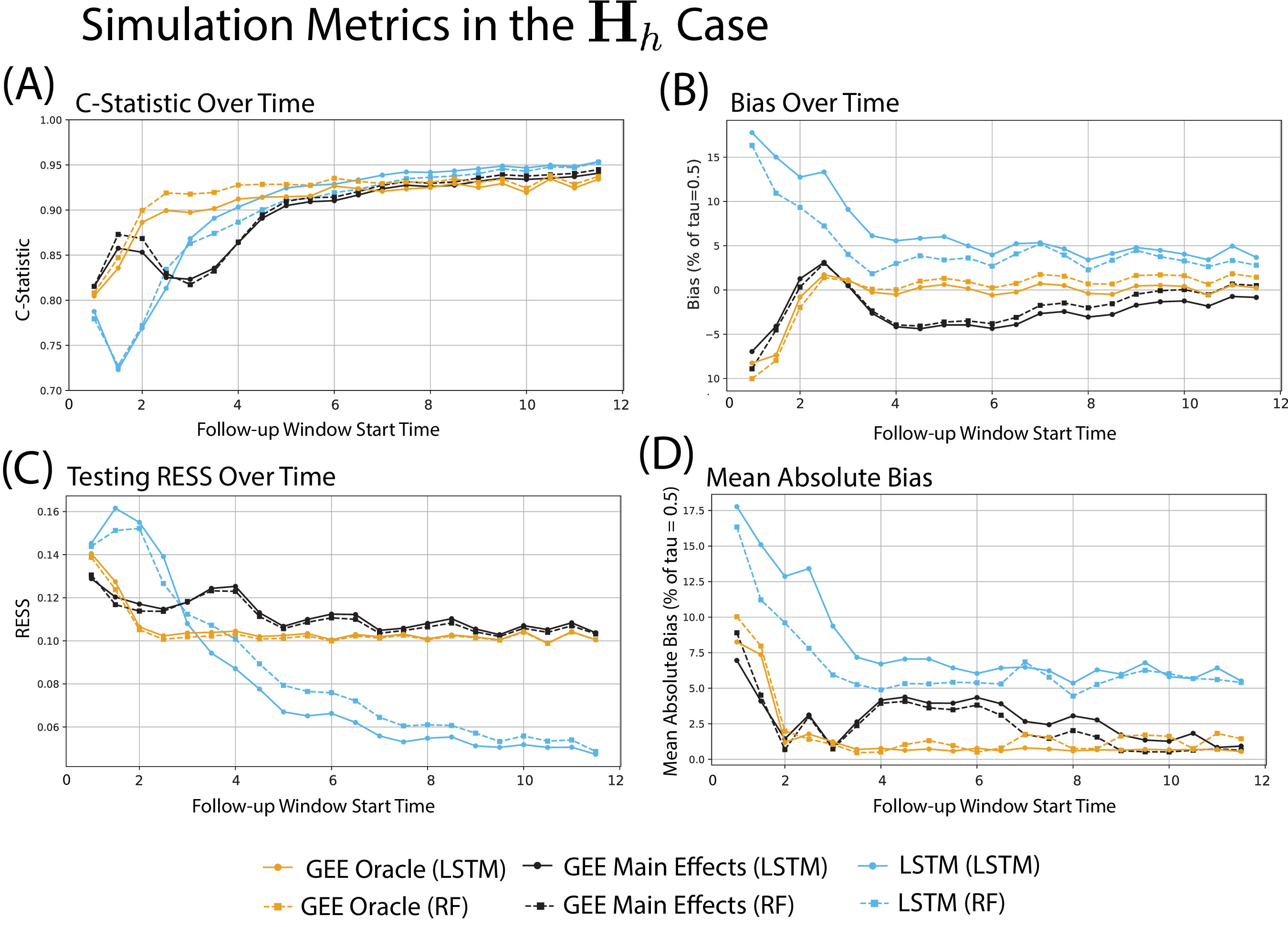}
    \caption{A panel graphic displaying the performance of the dynamic prediction models under consideration in Section \ref{s:simulation} for the $\bH_h$ setting. Counter clockwise from upper left is (A) c-statistic over time, (B) average bias over time, (C) testing root mean squared error over time, and (D) $n_{opt}$ over time. The method used to predict $E[\min\{T_i(t), \tau\}] = \Bar{\bZ}^\top \beta $ is listed first (yellow for GEE Oracle, black for GEE Main Effects, and blue for LSTM), while the weight model is listed in parenthetical (with LSTM weights graphed in a solid line, while random forest weights are dashed). All results are from $200$ replicates with $n = 2500$ each.\\
    Alt text: A panel graphic displaying simulation metrics for the dynamic prediction models in the high-quality history information setting.
    }
    \label{fig:last_two_metrics}
\end{figure}

Figures \ref{fig:only_hist_metrics} and \ref{fig:last_two_metrics} display dynamic prediction model results based on the testing dataset at each $t\in \mathcal{T}$. See Supplementary Materials Figures \ref{fig:supp:only_banded} and \ref{fig:supp:last_two_banded} for metrics with corresponding 95\% confidence bands. Figure \ref{fig:only_hist_metrics} reflects the scenario where only history variables collected during study follow-up are known ($\bH_\ell$ setting) and Figure \ref{fig:last_two_metrics} reflects the setting where history covariates are known at all times, $t\in \mathcal{T}$ ($\bH_h$ setting).
Panels A, B, C and D show the average C-statistic, percent bias, RESS results,  and percent absolute bias respectively, all in the testing set (data held out and not used in training). In each panel, GEE Oracle results are graphed in yellow, GEE Main Effects results are in black, and LSTM model results are in sky blue. Each dynamic prediction model is assessed with use of random forest or LSTM IPW weights, which are labeled in figure legends parenthetically after the model designation and graphed with a dashed line (RF weight), or a solid line (LSTM weight). 
%The same metrics are shown for the $\bH_h$ models in Figure \ref{fig:last_two_metrics}

The GEE Oracle model performs well in either scenario, of course, with performance metrics continuing to improve as on-study history covariates enrich over time. All models generally improve as history covariate information grows over time. Evaluating inverse weighting methods for the GEE Main Effects model and GEE Oracle model, both RF and LSTM inverse weights perform similarly in terms of the c-statistic, RESS and bias metrics. For GEE Oracle, at later time points, RF weights appear slightly less biased in $\bH_\ell$ settings, while LSTM weights are slightly less biased in $\bH_h$ settings. When considering GEE Main effects, at later points LSTM weights have slightly less bias for $\bH_\ell$ settings, and RF weights have slightly less bias for $\bH_h$ settings. Bias at early windows is attributed to the online prediction nature of fitting both inverse weights and restricted means. Lack of data at early windows causes machine learning model based fitted values for $P\{R_i(t)=1\}$ to be less reliable than fitted values at later windows (which are based on more data).

At later time points, the long short-term memory network outperforms the GEE Main Effects model in both $\bH_\ell$ and $\bH_h$ settings in terms of C-statistic and testing RESS, though LSTM-based models initially performed poorly at early follow-up windows. In the $\bH_\ell$ case, the LSTM with LSTM weights and RF weights both have superior c-statistics and minimal bias at later windows; at some time points, the LSTM-based restricted mean model c-statistics even outperform the c-statistic corresponding to the semiparametric oracle model. The LSTM model with LSTM weights performs well in all follow-up windows in terms of bias, with the least bias of all approaches in early follow-up windows; bias results for LSTM approaches over time were more variable than GEE approaches.

In the $\bH_h$ setting, the richer covariate history greatly improves performance of the GEE Main Effects approach relative to the $\bH_\ell$ setting, with higher c-statistics over time that are competitive with GEE oracle after midway through the study period. The LSTM based models initially perform poorly in terms of c-statistic, testing RESS, and bias. However, metrics rapidly improve over the first third of the study period, corresponding to a need for plenty of follow-up for the LSTM-based models. LSTM RESS for later time points is roughly half of that corresponding to the oracle model (Figure \ref{fig:last_two_metrics}, Panel C), with outstanding c-statistics relative to both semiparametric models. However, percent bias appears to stabilize around 5\% for the neural network architectures, while the oracle model stabilizes around 0, in a classic bias and variance trade-off.

In general, the selected architecture for the neural networks in simulation justifiably favored smaller RESS over smaller bias, as we selected architectures based on lowest testing loss (weighted squared errors). Excellent performance of LSTM-based prediction architectures in this setting echoes improvements seen in other neural network applications in machine learning literature. Specifically, use of embeddings allows for correlation within time and individuals to be modeled implicitly in the LSTM, while increased dimensions of $\eta(t)$, particularly with $\bH_h$ covariates, allows for complex emergent relationship between covariates and outcomes. There are no equivalent such customization in semiparametric modeling, though embeddings loosely resemble random effects modeling. 

\begin{table}[h]
\begin{tabular}{l|lccccc}
History & Weight  & Subj. Embed Dim. & Time Embed Dim.  & $h$  & No. LSTM layers  & $p_d$ \\
\hline 
 $\bH_\ell$ &LSTM  &1  &8  &8  &1  &0  \\
 &RF   &1  &8  &4  &1  &0  \\
 \hline
 $\bH_h$& LSTM  &1  &8  &64  &2  &0.2  \\
 &RF  &1  &8  &4  &1  &0.2 
\end{tabular}
\caption{A table displaying the hyperparameters selected for the LSTM models in Section \ref{s:simulation}. Hyperparameters were selected based on lowest testing loss in a rolling-origin grid search. Weight refers to the weighting strategy used in the neural network loss function, while ``Subj. Embed Dim." and ``Time Embed Dim." refer to the dimension of the vector that the network uses to learn time and subject characterizations. Hyperparameter $h$ refers to the dimension of the ``hidden state," or number of columns of the linear predictor $\eta$, where $\eta \in \mathbb{R}^{n(t) \times h}$. The number of LSTM layers refers to the number of LSTM architectures stacked on top of each other, and $p_d$ is the proportion of slopes and intercept randomly set to 0 during each epoch of training. For further details on these parameters, see Section \ref{s:nnet:lstmDesc}.}\label{tab:hyperparams}
\end{table}

\section{Application to Intern Health Study Data}
\label{s:application}
Medical residency, or the period of medical training after medical school, is known to be a stressful period of time that may lead to depression with a rate higher than in the general population \citep{guille2010ProspectiveIHS}, with negative effects on job satisfaction, physician retention and medical errors \citep{fang2022IHS}. It is of interest to identify opportune moments of potential interventions, e.g., delivered via mobile app prompts, to improve the mental health and general well-being of the training physicians \citep{necamp2020assessing, wang2023effectiveness}. We uses data obtained from the 2023 cohort of the Intern Health Study, a multi-center study, where interns self-reported mood on a scale of 1 to 10 (10 being best) daily. It is of clinical interest to dynamically predict periods when mood for individual interns falls 3 points below baseline mood scores. 

Covariates available for analysis include: age, gender, marital status, race, ethnicity, residency location and program, SEQ (a measure of adverse sexual experiences), Risky Families Questionnaire score (a measure of exposure to adverse conditions as a child), neuroticism at baseline, daily steps, sleep minutes, resting heart rate, mobile health device (Garmin, Apple Watch or FitBit), bedtime, wake time, if a participant has a history of depression, neuroticism or disability, any treatment for depression, substance use, and birth place. Incomplete baseline covariates are handled via a single imputation of data using random forest \citep{micePackage}. Missing mobile health variables after baseline are imputed using last observation carried forward. As there are no mood scores for individuals prior to study start time, history covariates are imputed using a random draw from an appropriate exponential distribution at $t_0$, and updated at later window start times. %Providing a flexible prediction model for interns based on covariates and mobile health variables would be of use to the medical profession.

The censored longitudinal data structure assumes the form $\mathcal{T} = \{0, 14, 28, \ldots, 350\}$ days with $\tau = 14$ days. As in Section \ref{s:simulation}, IPW weights are estimated via longitudinal random forest or an LSTM model at each time point. The inverse probability LSTM uses 2 layers, $h=70$, 2 dimensions for subject and time embeddings, and $p_d = 0.3$, with 2-fold cross-validation to select number of epochs to train.

We fit the following main effects model via GEE at each time point for dynamic prediction:
% % as.formula("pseudoval~ Age + Sex+GenderIdentity+Marital+ partnerEmploy+
%                              Neu0+depr0+SEQtot+EFE0+deprTreat0+
%                              steps+restingheartrate+Ethnicity+
%                              SpecialtyAAMC+
%                              depr0+tobacco0+alcohol0+deprAge+disabilityYN+
%                              device+sleepduration+
%                     last_event_1+avg_time_since_last_event+avg_cycle_duration+
%                     num_events_observed+avg_time_recovered+avg_time_depressed")
$$
E[\min\{T_i(t), \tau\}\mid \bZ_i(t)] = \beta^\top {\bZ}_i(t),
$$
where ${\bZ}_i(t)$ contains all covariates listed above, and derived history covariates. History covariates include the gap times between last recurrent event pairing, time since last mood score drop, average time spent in the risk population ($R_i(t) =1$), average time spent with lowered mood, and observed number of lowered mood events. Parameter estimates based on the GEE main effects model at $t = 350$ are displayed in Supplemental Materials \ref{s:supp:dataApp}. 

LSTM architectures for the neural network weights and the random forest weights are shown in Table \ref{tab:dataAppHyperparams}. The network selected for fitting a restricted mean model using LSTM weights is relatively simple, particularly when compared to the model selected for use with RF weights. The architecture corresponding to LSTM with RF weights has fewer embeddings for subject and time, but the hidden size (number of columns of $\eta$) is over five times that of the hidden size of the LSTM with LSTM weights. In addition to this complexity, the rolling-origin grid search algorithm indicated the use of three separate LSTM layers with drop out of $0.2$.
%, with categorical variables transformed into reference level indicators.

\begin{figure}
    \centering
    \includegraphics[width=.65\linewidth]{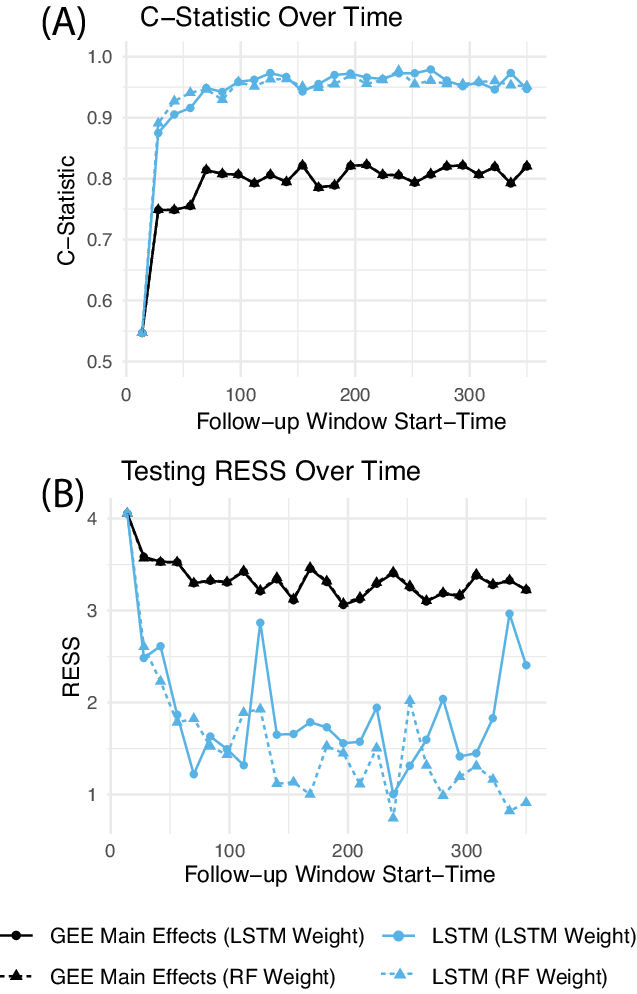}
    \caption{A panel graphic displaying results from predicting periods of low mood for medical interns. The censored longitudinal data structure takes the form $\tau=14$, $\mathcal{T} = \{0, 14, 28, \ldots, 350\}$.  Panel (A) displays c-statistic, while Panel (B) displays testing RESS. The main effects model fit via GEE is graphed in black, while the LSTM models are shown in sky blue. The method used to calculate IPW weights is listed in parenthetical after the model name.\\
    Alt text: A panel graphic displaying c-statistic, and testing RESS at each follow-up window start-time for both the LSTM method and the semiparametric method for the 2023 cohort of the IHS.}
    \label{fig:placeholder}
\end{figure}

Figure  \ref{fig:placeholder} shows predictive performance via C-statistics  over time and RESS in the testing cohort. The neural network algorithm had outstanding C-statistic performance, with a maximum of approximately 0.97 at day 266 or 238 (Panel A), depending on weighting scheme used. In comparison, the GEE Main Effects model C-statistic topped out at 0.823 on day 210. The manner of inverse weight estimation gave similar results in terms of C-statistics shown in Panel A. Panel B shows smaller RESS compared to the GEE Main Effects model. The LSTM model using LSTM inverse weights had the smallest RESS of all methods after approximately day 300 on study.

\begin{table}[h]
\begin{tabular}{l|lccccc}
History & Weight  & Subj. Embed Dim. & Time Embed Dim.  & $h$  & No. LSTM layers  & $p_d$ \\
\hline 
 $\bH_\ell$ &LSTM  &1  &8  &6  &1  &0  \\
 &RF   &1  &1  &32  &3  &0.2  
\end{tabular}
\caption{A table displaying the hyperparameters selected for the LSTM models in Section \ref{s:application}. Hyperparameters were selected based on lowest testing loss in a rolling-origin grid search at $t_3$. Weight refers to the weighting strategy used in the neural network loss function, while ``Subj. Embed Dim." and ``Time Embed Dim." refer to the dimension of the vector that the network uses to learn time and subject characterizations. Hyperparameter $h$ refers to the dimension of the ``hidden state," or number of columns of the linear predictor $\eta$, where $\eta \in \mathbb{R}^{n(t) \times h}$. The number of LSTM layers refers to the number of LSTM architectures stacked on top of each other, and $p_d$ is the proportion of slopes and intercept randomly set to 0 during each epoch of training.}\label{tab:dataAppHyperparams}
\end{table}

\section{Discussion}
\label{s:discussion}
In this manuscript we developed several dynamic prediction approaches for the  alternating recurrent event setting that leverage the power of popular modern machine learning methods, and discussed the relative strengths and weaknesses of these methods. Of particular importance is the care with which we introduced the long short-term memory neural network machinery, with an eye towards bridging the notational and lexical gap between statistical  and computer science practitioners. Strengths of neural networks (LSTMs included) and random forest approaches for building prediction tools in other settings have been well documented \citep[e.g.,][]{Chapter1, breiman2001random, hochreiter1997long, Mikolov2012embeddings, beck2024xlstmNeurIPS, goodfellow2016deep}. This strong performance is somewhat offset by the lack of parameter estimates with well described statistical behavior that
%, and more flexible machine learning algorithms lack the estimates of variability 
one may obtain from a parametric or semiparametric statistical modeling approach, such as the GEE approach specified in Section \ref{s:ipwgee}. Such parameter estimates are often helpful for understanding underlying associations between predictors and outcomes that generate further scientific hypotheses. 

% I feel like maybe this sentence should not be ihere? we weren't focused on reproducibility? Both computer science and statistical methods researchers have effectively adopted strategies to promote reproducible predictions, via setting a random seed, specifying architectures, ensuring generalizability ***list of things like blah, blah blah for neural networks, blah, blah, blah for random forest and blah, blah, blah for GEE. 

In Section \ref{s:simulation}, each method performed well in the alternating recurrent event setting, but the LSTM approaches truly shined as individual history data grew over time and was incorporated into these algorithms. If not for the parameter estimate information related to history covariates seen in GEE algorithm settings (where linear terms of history covariates were consistently the most statistically significant in Tables \ref{supp:tab:lstm_gee} and \ref{supp:tab:rf_gee}), the reason for such outstanding performance for LSTM methods would remain completely opaque. As it stands, the true functional form of these history covariates within the LSTM algorithm still remains unclear. While there have been strides made in extracting relative importance of covariates in deep learning paradigms (e.g., Shapley values \citep{lundberg2017unified}, or the field of interpretable machine learning), no off-the-shelf, easily programmable method exists for our proposed LSTM.

On a more practical note, the proposed analysis methods are agnostic to the definitions of primary and secondary recurrent events. For example, in Section \ref{s:simulation}, we modeled the primary event type, corresponding to $\mu^{(1)}$; however, analysis of the secondary event type may be of equal import. In that case, an analyst may exchange the definitions of primary and secondary events, and obtain two dynamic online prediction models, one for each type.

In conducting simulations, we often observed that small differences in network architecture could affect results substantially (data not shown). If the goal is to minimize an objective function, neural networks tend to obtain a local minimum that  produces high levels of discrimination. In simulation, this may coincide with some bias in the estimated predictions. In statistical methods literature, such a bias-precision tradeoff is carefully mapped out through statistical proofs of consistency, identifiability and limiting behavior before popular use commences. The merging of computing and statistical literature for prediction is growing quickly, with technical arguments on why machine learning methods work well lagging somewhat behind. Content experts in both fields are aware that hyperparameter selection is a key concern in such algorithms. In our early simulation attempts, we saw firsthand that LSTM methods are not robust to hyperparameter choices made at random; the identification of hyperparameters from the testing set is key to a well-behaved LSTM algorithm. 
%rucially, our analysis attempted to identify hyperparameters that led to minimal loss in the testing set; random selection of hyperparameters in simulation led to poor performance. As such, predictions from an architecture are robust to misspecification of the mean model (in that there is not a mean model), but not to misspecification of the hyperparameters. 
Further work in understanding exactly how slight changes of hyperparameters affect network performance in our setting requires an ablation study \citep{ablation2019}, a direction we hope to continue working towards.

\backmatter

%  This section is optional.  Here is where you will want to cite
%  grants, people who helped with the paper, etc.  But keep it short!

\section*{Acknowledgments}

We thank the investigators, coordinators, and research associates or assistants who worked on the Intern Health Study (PI: Dr. Srijan Sen; \href{https://clinicaltrials.gov/study/NCT07052357}{NCT07052357}), the study participants, and the organizations who supported recruitment efforts. We also thank Xingran Chen for generously sharing his expertise in the \texttt{torch} package, and general start-up tips. ZW is partly supported by a grant R01NR013658 from National Institute of Health, National Institute of Nursing Research.

\section*{Data Availability Statement}
The IHS data that support the findings in this paper are available upon reasonable request. The R code for reproducing the simulation and data analysis is accessible via\\ \url{http://github.com/AbigailLoe/p3}. 

%  If your paper refers to supplementary web material, then you MUST
%  include this section!!  See Instructions for Authors at the journal
%  website http://www.biometrics.tibs.org

% \section*{Supplementary Materials}

%  Here, we create the bibliographic entries manually, following the
%  journal style.  If you use this method or use natbib, PLEASE PAY
%  CAREFUL ATTENTION TO THE BIBLIOGRAPHIC STYLE IN A RECENT ISSUE OF
%  THE JOURNAL AND FOLLOW IT!  Failure to follow stylistic conventions
%  just lengthens the time spend copyediting your paper and hence its
%  position in the publication queue should it be accepted.

%  We greatly prefer that you incorporate the references for your
%  article into the body of the article as we have done here 
%  (you can use natbib or not as you choose) than use BiBTeX,
%  so that your article is self-contained in one file.
%  If you do use BiBTeX, please use the .bst file that comes with 
%  the distribution.
% \pagebreak

% \begin{thebibliography}{}
% \bibliographystyle{biom} 
% \bibliography{biomsample}
% \end{thebibliography}

% \appendix

\bibliographystyle{biom} 
\bibliography{mybib}

\appendix

%  To get the journal style of heading for an appendix, mimic the following.

% \section{}
% \subsection{Title of appendix}

% Put your short appendix here.  Remember, longer appendices are
% possible when presented as Supplementary Web Material.  Please 
% review and follow the journal policy for this material, available
% under Instructions for Authors at \texttt{http://www.biometrics.tibs.org}.

\label{lastpage}

\end{document}

\def\spacingset#1{\renewcommand{\baselinestretch}%
		{#1}\small\normalsize} \spacingset{1}
	
%%%%%%%%%%%%%%%%%%%%%%%%%%%%%%%%%%%%%%%%%%%%%%%%%%%%%%%%%%%%%%%%%%%%%%%%%%%%%%
%% below are commands for blinding/unblinding authors for review purposes
%%%%%%%%%%%%%%%%%%%%%%%%%%%%%%%%%%%%%%%%%%%%%%%%%%%%%%%%%%%%%%%%%%%%%%%%%%%%%%
\if0\blind
	{		
\title{\bf Supplementary Materials for \\``Dynamic Prediction of Alternating Recurrent Events via Neural Network"}
		\author[1,$^\ast$]{Abigail Loe}
		\author[2]{Susan Murray}
		\author[2]{Zhenke Wu}
        \affil[1]{Department of Mathematics, Statistics and Computer Science, Macalester College, Saint Paul, MN 55105, USA
        \\
        E-mail: $^\ast${\tt aloe@macalester.edu}
        }
		\affil[2]{Department of Biostatistics, University of Michigan, Ann Arbor, MI 48109, USA}
		
		\date{}
		\maketitle
} \fi
	
\if1\blind
	{
		\title{\bf Probability Weighted Pseudo-Observations for Alternating Recurrent Events}
		\author[]{}
		\date{}
		\maketitle
	} \fi
	
\vspace{-1cm}
%\begin{center}
%{\sf \textbf{Version}: 1\\
% \textbf{Compiled}: \date{\today} at \currenttime}
%\end{center}

\bigskip

%\noindent%

%\vfill

%\spacingset{1.45} 
% DON'T change the spacing!

\renewcommand{\thefigure}{S\arabic{figure}}
\renewcommand{\theequation}{S\arabic{equation}}
\renewcommand{\thetable}{S\arabic{table}}
\renewcommand{\thetheorem}{S\arabic{theorem}}
\renewcommand{\thesection}{A\arabic{section}}

\setcounter{figure}{0}
\setcounter{equation}{0}
\setcounter{table}{0}
\setcounter{theorem}{0}

\section{A Brief Review of Long Short-term Memory Architectures}

\subsection{Long Short-term Memory Neural Networks}\label{s:nnet:lstmDesc}

There are many variations of (recurrent) neural network architectures intended to accommodate longitudinal data. Here we review LSTMs as a point of entry into  more intricate network structures for more complex data largely for three reasons: (1) implementable software is available, (2) LSTMs are known to handle long-range time-dependencies \citep{hochreiter1996lstm}, and (3) much effort has been made by computer scientists to de-mystify their performance.

%As the field progresses, this stage of our analysis approach can be replaced with more modern versions of deep learners, such as transformers or gated recurrent units.  

\begin{figure}
    \centering
    \includegraphics[width = .75\textwidth]{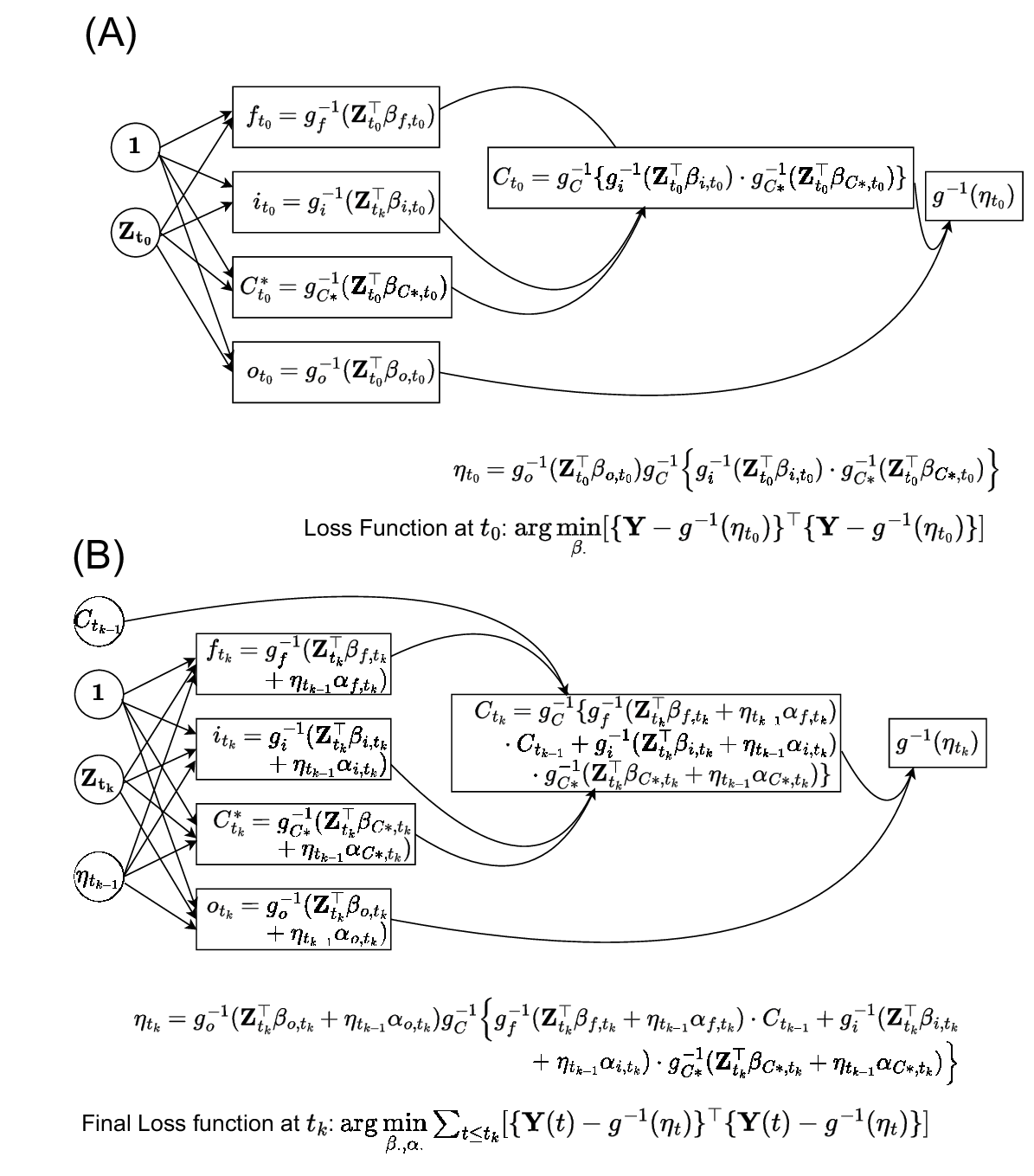}
    \caption{A panel graphic displaying the network architecture termed ``long short-term memory," or LSTM at time $t_k$. Panel (A) displays the functionals used in training an LSTM at time $t_0$, while Panel (B) displays the recursive architecture for later $t \in \mathcal{T}$. Software package \texttt{torch} allows user-specification of  links $g_f, g_i, g_{C*}, g_o$ and $g_C$ but defaults to $g_f, g_i, g_{C*}$ and $g_o$ as the log-link and $g_C$ and $g_{C*}$ as the inverse hyperbolic tangent; thus $g^{-1}_{f, i, C*, o} (x) = \frac{\exp(x)}{1+ \exp(x)}$, and $g_{C, C*}^{-1}(x) = \tanh(x)$. In Section \ref{s:simulation}, we used defaults from the \texttt{torch} package in Python. Node outputs from $C^*_{t_k}$ and $i_{t_k}$ are combined together via a dot product, as are $f_{t_k}$ and $C_{t_{k-1}}$, and $C_{t_k}$ and $o_{t_k}$. Not all covariates are used in all nodes in the first layer; covariate $C_{t_{k-1}}$ is only used in the second layer, while $o_{t_k}$ is calculated in the first layer, before final combination in the third. At each time point, inputs that depend on the algorithm at the prior time point itself are included ($C_{t_{k-1}}$ and $\eta_{t_{k-1}}$). At $t_0$, these quantities default to $0$.
    In Panel (B), while ultimate output of interest is $g^{-1}(\eta_{t_k})$, loss from the previous time point factors in to the modeling of $\eta_{t_k}$.\\
    Alt. text: A panel graphic displaying the recursive structure of the long short-term memory network architecture. The base case is in Panel A, while the recursive step is in Panel B.}
    \label{fig:lstm}
\end{figure}

A bare-bones LSTM for a single time-point, $t_0,$ is depicted in Figure \ref{fig:lstm}A, with two hidden layers, an intercept (circled $\mathbf{1}$) and a single covariate (circled $\bZ_{t_0}$). Figure \ref{fig:lstm}B depicts a generalization of the LSTM network at time $t_k$; taken together, Panels A and B recursively define the architecture termed LSTM. Unlike the networks depicted in Figure \ref{fig:basicnet} where each covariate and node is connected to subsequent layers, the $o_{t_k}$ node skips from the first hidden layer to the calculation of $\eta_{t_0}$. This style of neural network architecture is designated as not fully connected in computer science literature. 
%the layers are not fully connected. 
%For example, output from node $o_{t_k}$ is calculated in the first hidden layer, but not used until calculating $\eta_{t_k}$ in the final layer.

LSTMs have the advantage of connecting architecture designed for particular follow-up times, $t\in\{t_0,\ldots,t_b\}$, into an overall super-architecture. Together, panels \ref{fig:lstm}A and \ref{fig:lstm}B recursively define a full LSTM super-architecture, where nodes for $C_{t_{k-1}}$ and $\eta_{t_{k-1}}$ connect the architecture for $t_{k-1}$ to that for $t_k, k=0,\ldots, b$. As in \citet{loe2026neuralnetworkslinearregression}, $\bZ_{t_0}^\top \beta$ indicates the use of a design matrix, where $\beta$ is assumed to contain the intercept term $\beta_0$.
%using the output from hidden nodes at previous time points, functionals $C_{t_{k-1}}$ and $\eta_{t_{k-1}}$, in modeling $\eta_{t_k}$. 
The contributions of $C_{t_{k-1}}$ and $\eta_{t_{k-1}}$ are traditionally weighted by other nodes in the $t_k$-level architecture (a strategy in deep learning known as ``gating"). Gating connects longitudinal information between architectures, while avoiding a well-known ``vanishing gradient" problem (see, for further details, \citet{bengio1994learning}).

\section{Neural Network Specifications}\label{s:LSTMspecifics}
Training, validation and testing cohorts in the real-time LSTM prediction setting require in-depth explanation. Here we also wish to use grid search methods to optimize, for instance, the number of training iterations selected from the set, $\{1,\ldots,n_a\},$ and the learning rate, $\triangledown$ selected from a size $m$ grid of candidate learning rates, $\{\triangledown_1, \triangledown_2\ldots, \triangledown_m\}, $ where the optimal values of these hyperparameters may differ for dynamic predictions at the different follow-up window start times, $t_k\in \mathcal{T}.$ While in the following, we describe a grid search for two hyperparameters, all hyperparameters may be selected via such an algorithm, though a large number of points in a grid increases significantly increases computation time.
%(though all hyperparameters may be selected via a grid search)
Suppose our current focus is finding optimal hyperparameters for the follow-up window starting at $t_k$. At all $t \in \mathcal{T}$ the network described in Figure \ref{fig:lstm} may be fit. However, in constructing a training and validation set for hyperparameter selection, we require at least one window with complete follow-up for each of the training and validation data cohorts for use in the more sophisticated rolling-origin algorithm (proposed by \citet{armstrong1972comparative} and more concisely summarized by \citet{tashman2000out}.) This results in data from (at least) $t_0$ and $t_1$ forming a model warm-up period, where predictions are less trustworthy due to a lack of hyperparameter validation.
%For selection of learning rate and number of training epochs, first, we specify a size $m$ grid of candidate learning rates, $\{\triangledown_1, \triangledown_2\ldots \triangledown_m\}$. 
In utilizing the rolling-origin algorithm for the optimal learning rate and number of training epochs, for each candidate learning rate, $\triangledown_\ell, \ell=1,\ldots,m$, we first train on data observed up to $t_{k-2}$, using observed data at $t_{k-1}$ as a validation set, assuming $t_{k-1}<t_k-\tau$ (if $t_{k-1}> t_k-\tau$, $t_{k-1}$ and $t_{k-2}$ revert to the two most recent window start times with complete follow-up). The testing cohort becomes observed data at $t_k$. Behavior of the loss function in the training set is recorded after each training iteration completes for each learning rate, so that the (joint) optimal number of training iterations, $n_{opt}(t_k)$, and the optimal learning rate, $\triangledown_{opt}(t_k),$ can be selected.   Corresponding loss function performance in the validation cohort using the same set of candidate values for these hyperparameters are then recorded. 

Iterative algorithms are known to continue decreasing loss in the training data, thereby attaining a local minimum and overfitting to training data. As a result, the use of training, validating and testing with ``patience," or a threshold of iterations where validation loss is allowed to increase before stopping training early is a common practice, and one we employ. For the censored longitudinal dataset, we perform a custom grid search to determine the number of epochs to train to, and learning rate as described in \citet{loe2026neuralnetworkslinearregression}, at each time point using the rolling-origin algorithm. We select learning rate and number of epochs to train with at each $t\in \mathcal{T}$ corresponding to the pairing that yields minimal loss in the validation set. A second network is then trained with $\triangledown_{opt}(t_k)$ and $n_{opt}(t_k)$ on data from $\{t_0, t_1, \ldots t_{k-2}, t_{k-1}\}$. Network performance for dynamic prediction is tested using data observed at $t_k$. For a depiction of this algorithm, see Figure \ref{fig:rolling_origin}. 

This custom grid-search in combination with the rolling-origin algorithm has three attractive properties: (1) learning rate is generally considered one of the most important hyperparameters to tune \citep{goodfellow2016deep}, (2) the splitting algorithm maps neatly onto the censored longitudinal data, and %(3) as long as $\tau \leq t_{k}-t_{k-1}$ for all $k$, training, testing and validation data remain separate from each other, and  **** IF THERE IS NO OVERLAP, IE THAT THE \tau \leq t_{k}-t_{k-1}, THIS MAPS NICELY. IN CASE WE GET ASKED ABOUT OVERLAP HERE. *****
(3) this algorithm allows for the use of learned subject specific embeddings to model correlation within individuals.

\section{History Covariates Used in Simulation}\label{s:supp:histCov}
%"only.avg.risk.time","only.most.recent.discharge","only.length.of.last.hosp","only.avg.hosp.time","only.length.of.last.at.risk","only.most.recent.event.cal.time","only.n.events.seen"
We derive seven history covariates of the type described in \citet{Chapter1}. Covariates include: (1) mean time in the at-risk population, (2) time since last return to risk-population, (3) length of last removal from risk population, (4) average time removed from the risk population, (5) length of last fully observed time at-risk, (6) most recent observed event time, and (7) the number of alternating recurrent events observed. In calculating these covariates for $\bH_\ell$ settings, we use a single imputation for time $t_0$, and continue to use the imputation until at least one pair of alternating recurrent events has been observed on study time. These covariates are used in the restricted mean LSTM models.

We also include indicators of lagged at-risk status; that is, at $t_k$ we may define earlier at risk status for $t_{k-\ell}, \ell=1,\ldots,(k-1)$ via the indicator, $\mathcal{L}_i(t_k, \ell)= R_i(t_{k-\ell}).$ Collectively, this past recurrent event status history  as measured by $\mathcal{L}_i(t_k, \ell)$ summarizes the missingness pattern of our data \citep[see, for example,][]{RodRubin, TchetgenTchetgen, chen2026predictionbasedinferenceelectronichealth}.

In incorporating history covariates in the longitudinal random forest of the type from \citet{SecondProject}, an analyst selects variables to always include in the construction of each random tree. For $\bH_\ell$ settings, where only history observed on study time is recorded, and history covariates at $t_0$ are imputed, an analyst may take a pragmatic approach, and only force splits on history covariates after enough follow-up has commenced. In our simulations, required splits on history covariates for random forest weight estimation occurred only after $t_5$.

\section{Oracle Model}\label{s:supp:oracle}
\begin{figure}
    \centering
    \includegraphics[width=\textwidth]{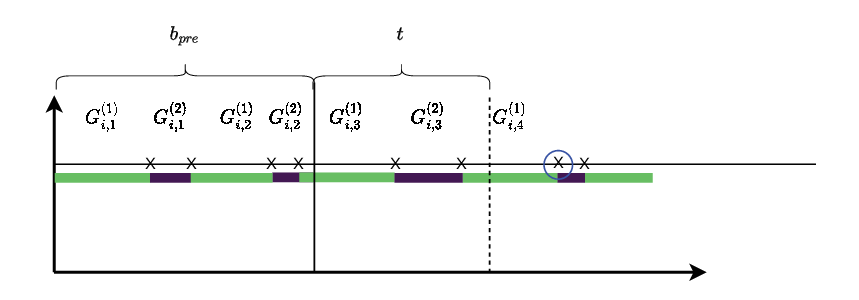}
    \caption{A graphic displaying the generating model for simulation, where $b_{pre}$ denotes burn-in, $t$ is window start-time beyond burn-in, and $G_{i,j}^{(k)}$ denote the correlated multivariate normal gap-times. The primary alternating recurrent event of interest that forms $T_i(t)$ is circled in blue, while primary event gap times are displayed in green, and secondary event times are displayed in purple.}
    \label{fig:supp:oracle_dist}
\end{figure}

Here we provide details for those interested in the derivation of the oracle model for $\min\{T_i(t), \tau\}$. Figure \ref{fig:supp:oracle_dist} displays a sample individual with burn-in length $b_{pre}$ and follow-up window start time $t$. Events are denoted by ``X``, with primary events displayed in green, and secondary events in purple. The $T_{i,j}^{(1)}$ that corresponds to the same event in creating $T_i(t)$ is circled in blue. We may re-write $T_{i,j}^{(1)} $ as $ G_{i,j}^{(1)}+\sum_{k<j}G_{i,k}^{(1)}+G_{i,k}^{2}$, where the marginal distribution of each $G_{i,j}^{(1)}$ is normal, with mean $\mu_i^{(1)}$ and variance $\sigma^2$, and $G_{i,j}^{(2)}$ is normal, with mean $\mu_i^{(2)}$ and variance $\sigma^2$. Re-arranging and subtracting $t$, we obtain $T_i(t) \sim \text{Normal}\big \{N_i^{(1)}(t^-)\mu_i^{(1)}+N_i^{(2)}(t^-)\mu_i^{(2)} - (t+b_{pre}), \Tilde{\sigma}^2 \big \},$ where $N_i^{(k)}(t^-)= \sum_j I(T_{i,j}^{(k)}<t + b_{pre})$ is the number of events of type $k$  experienced by subject $i$ prior to time $t$ including those accumulated during the burn-in period of duration, $b_{pre}$. 

\begin{figure}
    \centering
    \includegraphics[width=\textwidth]{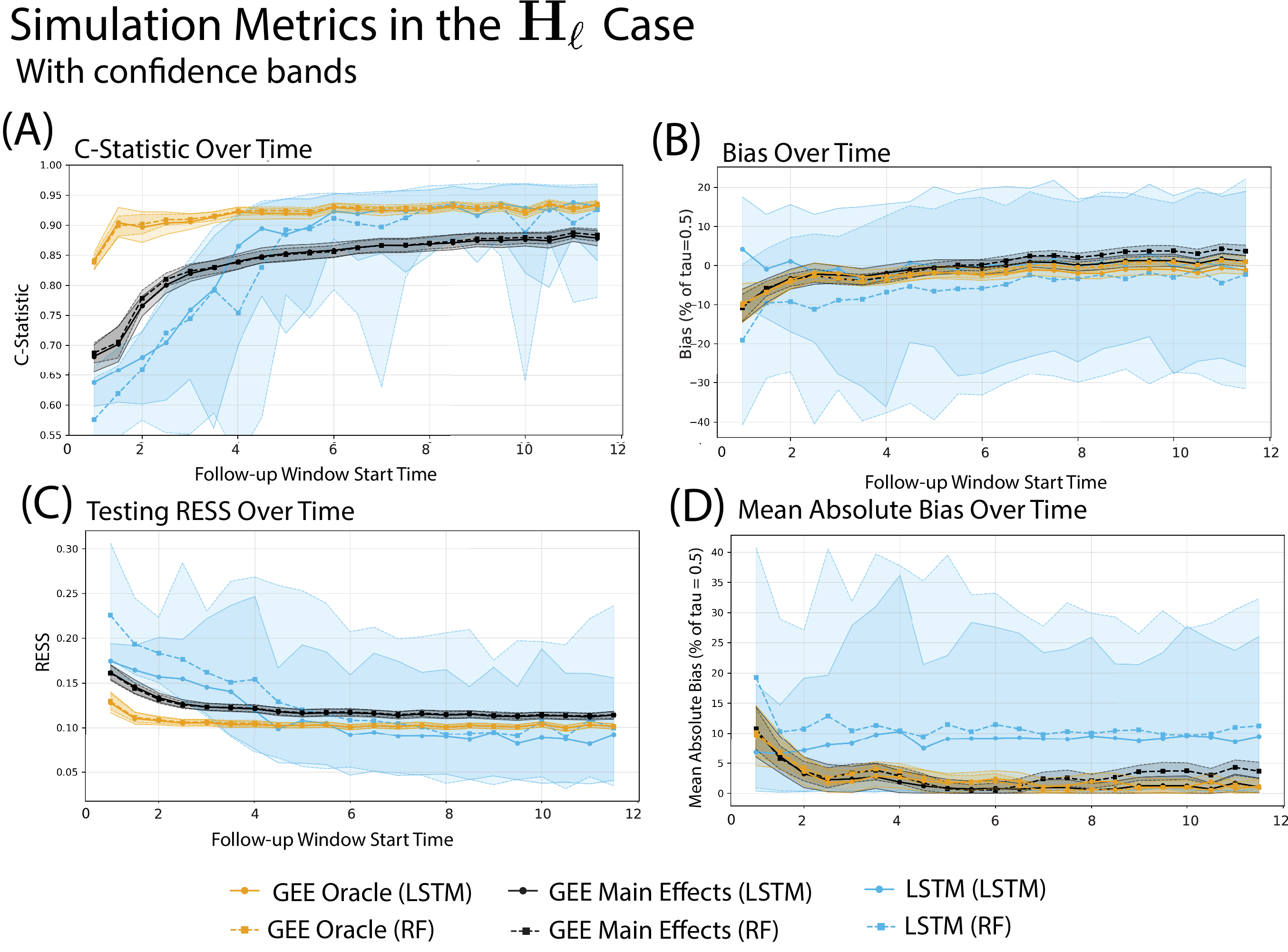}
    \caption{A figure displaying simulation metrics, with 95\% confidence bands showing the distribution of the metric under consideration.}
    \label{fig:supp:only_banded}
\end{figure}

\begin{figure}
    \centering
    \includegraphics[width=\textwidth]{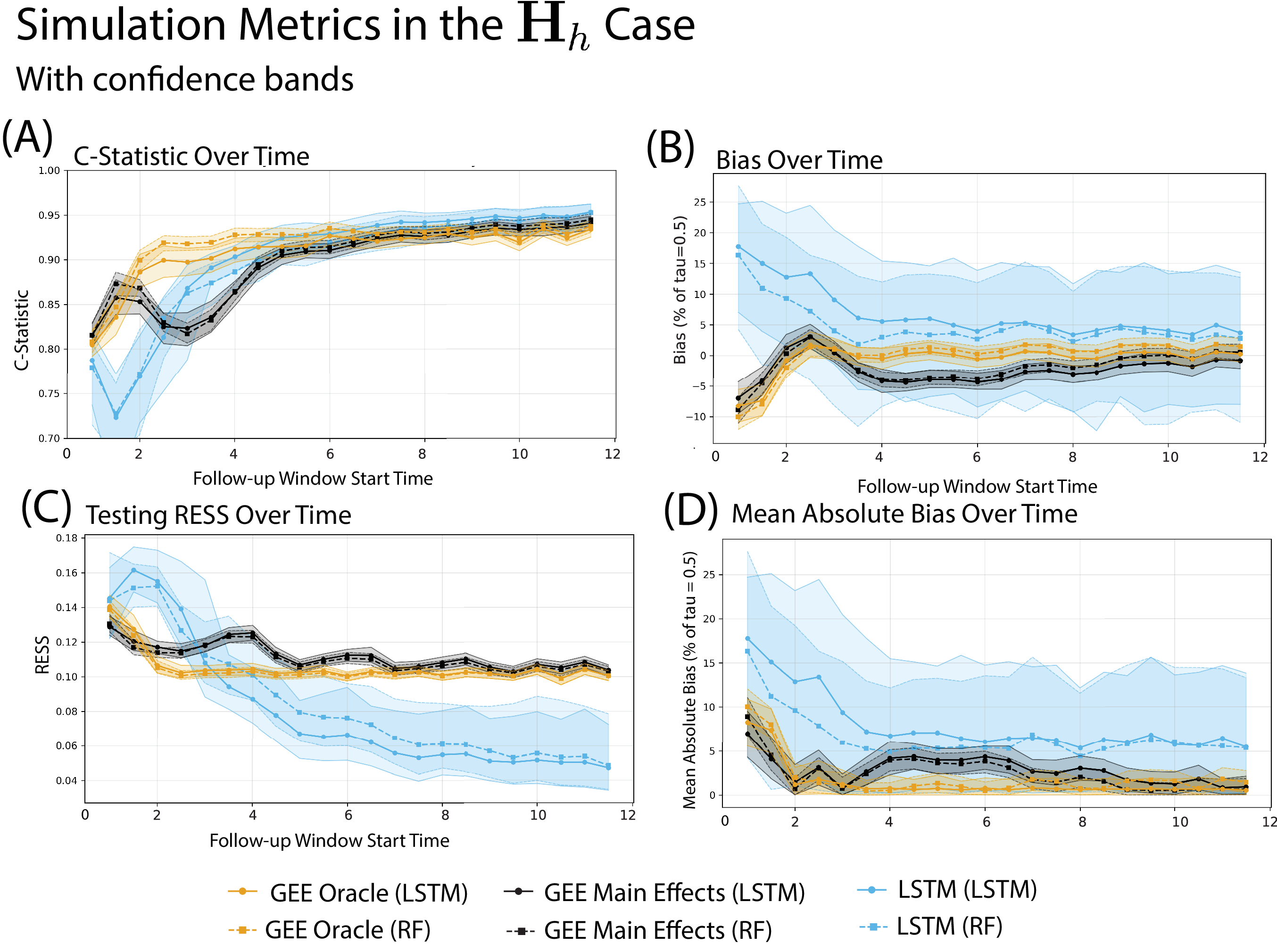}
    \caption{A figure displaying simulation metrics, with 95\% confidence bands showing the distribution of the metric under consideration.}
    \label{fig:supp:last_two_banded}
\end{figure}

\section{Data Application}\label{s:supp:dataApp}

\begin{longtable}{l|r|r|r|r}\caption{The point estimates with corresponding standard errors for the main effects model at time $t=350$ for the IHS data with LSTM weights. The reference levels for marital status is single, device reference is ``unknown", tobacco use at baseline is ``none", and alcohol use is ``none." The reference level for sex is male. History covariates of the type $\bH_\ell$ are denoted with a $^\dagger$.}\label{supp:tab:lstm_gee}\\
% \hline
Covariate & Estimate & SE & Z & P-value\\
\hline
\endfirsthead
%\hline
Covariate & Estimate & SE & Z & P-value\\
\hline
\endhead
Intercept & 11.6873 & 0.4730 & 610.4776 & $<$\textbf{0.0001}\\
%\hline
Age & 0.0099 & 0.0092 & 1.1539 & 0.2827\\
%\hline
Sex & -0.0583 & 0.0607 & 0.9226 & 0.3368\\
%\hline
Marital Status: Married & -0.0043 & 0.0987 & 0.0019 & 0.9655\\
%\hline
Marital Status: Separated/Divorced & -0.8771 & 0.6560 & 1.7880 & 0.1812\\
%\hline
Marital Status: Not in Committed Relationship & 0.0403 & 0.0954 & 0.1788 & 0.6724\\
%\hline
Marital Status: In a Committed Relationship & -0.0122 & 0.0940 & 0.0167 & 0.8971\\
%\hline
Neuroticism Score at Baseline & -0.0054 & 0.0034 & 2.5320 & 0.1116\\
%\hline
Sexual Experiences Questionnaire Score & 0.0051 & 0.0088 & 0.3286 & 0.5665\\
%\hline
Early Environment Exposures at Baseline & 0.0075 & 0.0027 & 7.5134 & \textbf{0.0061}\\
%\hline
Depression Treatment at Baseline & 0.0395 & 0.0751 & 0.2763 & 0.5991\\
%\hline
Standardized Daily Steps & -0.0510 & 0.0253 & 4.0439 & \textbf{0.0443}\\
%\hline
Resting Heart Rate & -0.0037 & 0.0013 & 8.1701 & \textbf{0.0043}\\
%\hline
Specialty: Child Neurology & 0.3999 & 0.3374 & 1.4047 & 0.2359\\
%\hline
Specialty: Dermatology & -0.3200 & 0.5579 & 0.3290 & 0.5663\\
%\hline
Specialty: Emergency Medicine & 0.1041 & 0.1313 & 0.6285 & 0.4279\\
%\hline
Specialty: Family Medicine & 0.1224 & 0.1453 & 0.7098 & 0.3995\\
%\hline
Specialty: General Surgery & -0.0667 & 0.1524 & 0.1917 & 0.6615\\
%\hline
Specialty: Internal Medicine & 0.3034 & 0.1170 & 6.7249 & \textbf{0.0095}\\
%\hline
Specialty: Neurological Surgery & 0.2012 & 0.3173 & 0.4020 & 0.5260\\
%\hline
Specialty: Neurology & 0.4683 & 0.1883 & 6.1819 & 0.0129\\
%\hline
Specialty: Obstetrics And Gynecology & -0.0459 & 0.1696 & 0.0734 & 0.7865\\
%\hline
Specialty: Ophthalmology & -0.2862 & 0.3056 & 0.8766 & 0.3491\\
%\hline
Specialty: Orthopedic Surgery & 0.1094 & 0.2204 & 0.2464 & 0.6196\\
%\hline
Specialty: Otolaryngology & 0.1096 & 0.2705 & 0.1641 & 0.6854\\
%\hline
Specialty: Pathology & -0.2980 & 0.2573 & 1.3416 & 0.2468\\
%\hline
Specialty: Pediatrics & 0.3517 & 0.1330 & 6.9919 & \textbf{0.0082}\\
%\hline
Specialty: Physical Medicine & -0.2590 & 0.3647 & 0.5045 & 0.4775\\
%\hline
Specialty: Plastic Surgery & -0.9375 & 0.5966 & 2.4692 & 0.1161\\
%\hline
Specialty: Psychiatry & 0.1332 & 0.1449 & 0.8445 & 0.3581\\
%\hline
Specialty: Radiology & -0.1434 & 0.3688 & 0.1512 & 0.6974\\
%\hline
Specialty: Surgery - General & 1.2243 & 0.4464 & 7.5219 & \textbf{0.0061}\\
%\hline
Specialty: Thoracic Surgery - Integrated & -0.5654 & 0.4566 & 1.5334 & 0.2156\\
%\hline
Specialty: Transitional Year & 0.2116 & 0.1625 & 1.6950 & 0.1929\\
%\hline
Specialty: Triple Board & 1.9607 & 1.3371 & 2.1501 & 0.1426\\
%\hline
Specialty: Urology & -0.2539 & 0.2504 & 1.0275 & 0.3108\\
%\hline
Specialty: Vascular Surgery & 0.3300 & 0.3852 & 0.7341 & 0.3916\\
%\hline
Depression at Baseline & -0.2191 & 0.1537 & 2.0320 & 0.1540\\
%\hline
Tobacco Use at Baseline: At least some & -0.0382 & 0.0906 & 0.1780 & 0.6731\\
%\hline
Alcohol: Low & -0.0127 & 0.0898 & 0.0200 & 0.8875\\
%\hline
Alcohol: High & 0.0536 & 0.0863 & 0.3860 & 0.5344\\
%\hline
Depression Age & 0.0003 & 0.0068 & 0.0024 & 0.9610\\
%\hline
Disability (Yes) & -0.1752 & 0.0216 & 65.6693 & $<$\textbf{0.0001}\\
%\hline
Device: Apple Watch & 0.3150 & 0.3445 & 0.8361 & 0.3605\\
%\hline
Device: Fitbit & 0.3509 & 0.3512 & 0.9980 & 0.3178\\
%\hline
Device: Garmin & 0.1696 & 0.3648 & 0.2161 & 0.6420\\
%\hline
Sleep Duration in Minutes & -0.0004 & 0.0001 & 7.5397 & \textbf{0.0060}\\
%\hline
Time of Since Most Recent Event$^\dagger$ & -0.0094 & 0.0009 & 106.9214 & $<$\textbf{0.0001}\\
%\hline
Average Time Since Last Event$^\dagger$ & 0.0199 & 0.0010 & 408.0342 & $<$\textbf{0.0001}\\
%\hline
Average Cycle Duration$^\dagger$ & -0.0138 & 0.0014 & 93.7753 & $<$\textbf{0.0001}\\
%\hline
Number Of Events Observed$^\dagger$ & -0.2715 & 0.0086 & 987.7060 & $<$\textbf{0.0001}\\
%\hline
Average Time Recovered$^\dagger$ & 0.0128 & 0.0007 & 384.2739 & $<$\textbf{0.0001}\\
%\hline
Average Time Depressed$^\dagger$ & 0.0168 & 0.0050 & 11.2241 & \textbf{0.0008}\\
%\hline
\end{longtable}

\begin{longtable}{l|r|r|r|r}\caption{The point estimates with corresponding standard errors for the main effects model at time $t=350$ for the IHS data with RF weights. The reference levels for marital status is single, device reference is ``unknown", tobacco use at baseline is ``none", and alcohol use is ``none." The reference level for sex is male. History covariates of the type $\bH_\ell$ are denoted with a $^\dagger$.}\label{supp:tab:rf_gee}\\
% \hline
Covariate & Estimate & SE & Z & P-value\\
\hline
\endfirsthead
%\hline
Covariate & Estimate & SE & Z & P-value\\
\hline
\endhead
Intercept & 11.6871 & 0.4866 & 576.7915 & $<$\textbf{0.0001}\\
%\hline
Age & 0.0111 & 0.0094 & 1.4070 & 0.2356\\
%\hline
Sex & -0.0567 & 0.0627 & 0.8198 & 0.3653\\
%\hline
Marital Status: Married & 0.0008 & 0.1004 & 0.0001 & 0.9940\\
%\hline
Marital Status: Separated/Divorced & -0.9118 & 0.7043 & 1.6764 & 0.1954\\
%\hline
Marital Status: Not in Committed Relationship & 0.0346 & 0.0969 & 0.1275 & 0.7210\\
%\hline
Marital Status: In a Committed Relationship & -0.0068 & 0.0958 & 0.0051 & 0.9431\\
%\hline
Neuroticism Score at Baseline & -0.0069 & 0.0036 & 3.7278 & 0.0535\\
%\hline
Sexual Experiences Questionnaire Score & 0.0046 & 0.0090 & 0.2585 & 0.6111\\
%\hline
Early Environment Exposures at Baseline & 0.0085 & 0.0028 & 9.2391 & \textbf{0.0024}\\
%\hline
Depression Treatment at Baseline & 0.0402 & 0.0775 & 0.2690 & 0.6040\\
%\hline
Standardized Daily Steps & -0.0524 & 0.0258 & 4.1166 & \textbf{0.0425}\\
%\hline
Resting Heart Rate & -0.0040 & 0.0013 & 9.0606 & \textbf{0.0026}\\
%\hline
Specialty: Child Neurology & 0.5175 & 0.3518 & 2.1641 & 0.1413\\
%\hline
Specialty: Dermatology & -0.2775 & 0.5460 & 0.2583 & 0.6113\\
%\hline
Specialty: Emergency Medicine & 0.1201 & 0.1358 & 0.7822 & 0.3765\\
%\hline
Specialty: Family Medicine & 0.1489 & 0.1493 & 0.9950 & 0.3185\\
%\hline
Specialty: General Surgery & -0.0826 & 0.1594 & 0.2686 & 0.6043\\
%\hline
Specialty: Internal Medicine & 0.3327 & 0.1221 & 7.4234 & \textbf{0.0064}\\
%\hline
Specialty: Neurological Surgery & 0.2930 & 0.3515 & 0.6949 & 0.4045\\
%\hline
Specialty: Neurology & 0.4508 & 0.1960 & 5.2897 & \textbf{0.0215}\\
%\hline
Specialty: Obstetrics And Gynecology & -0.0042 & 0.1733 & 0.0006 & 0.9808\\
%\hline
Specialty: Ophthalmology & -0.1588 & 0.3151 & 0.2541 & 0.6142\\
%\hline
Specialty: Orthopedic Surgery & 0.1915 & 0.2229 & 0.7375 & 0.3905\\
%\hline
Specialty: Otolaryngology & -0.0587 & 0.3019 & 0.0378 & 0.8459\\
%\hline
Specialty: Pathology & -0.3031 & 0.2655 & 1.3037 & 0.2535\\
%\hline
Specialty: Pediatrics & 0.3804 & 0.1374 & 7.6700 & \textbf{0.0056}\\
%\hline
Specialty: Physical Medicine & -0.9196 & 0.4488 & 4.1983 & \textbf{0.0405}\\
%\hline
Specialty: Plastic Surgery & -1.2734 & 0.6796 & 3.5109 & 0.0610\\
%\hline
Specialty: Psychiatry & 0.1615 & 0.1491 & 1.1738 & 0.2786\\
%\hline
Specialty: Radiology & -0.1946 & 0.3808 & 0.2612 & 0.6093\\
%\hline
Specialty: Surgery - General & 1.2241 & 0.4545 & 7.2542 & \textbf{0.0071}\\
%\hline
Specialty: Thoracic Surgery - Integrated & -0.6304 & 0.4358 & 2.0923 & 0.1480\\
%\hline
Specialty: Transitional Year & 0.1749 & 0.1694 & 1.0661 & 0.3018\\
%\hline
Specialty: Triple Board & 1.9272 & 1.3967 & 1.9041 & 0.1676\\
%\hline
Specialty: Urology & -0.2547 & 0.2575 & 0.9786 & 0.3225\\
%\hline
Specialty: Vascular Surgery & 0.3604 & 0.3839 & 0.8811 & 0.3479\\
%\hline
Depression at Baseline & -0.2088 & 0.1577 & 1.7524 & 0.1856\\
%\hline
Tobacco Use at Baseline: At least some & -0.0525 & 0.0941 & 0.3110 & 0.5770\\
%\hline
Alcohol: Low & -0.0124 & 0.0936 & 0.0175 & 0.8949\\
%\hline
Alcohol: High & 0.0659 & 0.0898 & 0.5378 & 0.4633\\
%\hline
Depression Age & -0.0008 & 0.0069 & 0.0124 & 0.9115\\
%\hline
Disability (Yes) & -0.1817 & 0.0220 & 67.9450 & $<$\textbf{0.0001}\\
%\hline
Device: Apple Watch & 0.2810 & 0.3570 & 0.6193 & 0.4313\\
%\hline
Device: Fitbit & 0.3334 & 0.3641 & 0.8381 & 0.3599\\
%\hline
Device: Garmin & 0.1256 & 0.3767 & 0.1111 & 0.7389\\
%\hline
Sleep Duration in Minutes & -0.0004 & 0.0001 & 9.2393 & \textbf{0.0024}\\
%\hline
Time of Since Most Recent Event$^\dagger$ & -0.0093 & 0.0010 & 94.6185 & $<$\textbf{0.0001}\\
%\hline
Average Time Since Last Event$^\dagger$ & 0.0200 & 0.0010 & 375.7507 & $<$\textbf{0.0001}\\
%\hline
Average Cycle Duration$^\dagger$ & -0.0135 & 0.0015 & 82.3069 & $<$\textbf{0.0001}\\
%\hline
Number Of Events Observed$^\dagger$ & -0.2712 & 0.0086 & 989.0799 & $<$\textbf{0.0001}\\
%\hline
Average Time Recovered$^\dagger$ & 0.0131 & 0.0007 & 396.4536 & $<$\textbf{0.0001}\\
%\hline
Average Time Depressed$^\dagger$ & 0.0127 & 0.0053 & 5.6517 & \textbf{0.0174}\\
%\hline
\end{longtable}

\bibliography{mybib}
\bibliographystyle{apalike}